\documentclass[11pt,twoside]{article}

\usepackage[utf8]{inputenc}
\usepackage{epsfig,amsfonts,color}
\usepackage{amsmath, bbm, mathabx}
\usepackage[normalem]{ulem}
\usepackage{amssymb, palatino, geometry,url}
\usepackage[colorlinks=true,linkcolor=blue,citecolor=blue,urlcolor=blue]{hyperref}
\usepackage{subcaption}
\usepackage{multirow}

\usepackage{fancyhdr}
\newcommand{\amin}{\mathop{\mbox{argmin}}}

\usepackage{lineno}
\usepackage{float}
\usepackage[section]{placeins}
\usepackage[table]{xcolor}

\title{PL-KKT-hPINN: Enforcing Nonlinear Equality Constraints on Neural Networks via Piecewise-Linear Projection}
\author{Fateme Mohammad Mohammadi$^1$, Hector Budman$^1$, and Joshua L. Pulsipher$^1$\thanks{Corresponding Author: pulsipher@uwaterloo.ca}\\
    {\small $^1$Department of Chemical Engineering}\\
	{\small \;University of Waterloo, 200 University Ave W, Waterloo, ON N2L 3G1, Canada}}
\date{}
\usepackage{indentfirst}
\usepackage[version=4]{mhchem}
\usepackage{bm}
\usepackage[font=footnotesize,labelfont=bf]{caption}

\begin{document}

\maketitle

\begin{abstract}
While physics-informed neural networks (PINNs) have shown strong potential for process modeling, physical equations are only enforced as soft constraints during training, and thus, they do not guarantee constraint satisfaction at inference. We propose a framework, called piecewise-linear Karush--Kuhn--Tucker hard-constrained PINNs (PL-KKT-hPINNs), that strictly enforces nonlinear equality constraints through piecewise-linear projection. This extends the KKT-hPINN framewor, which exactly enforces linear equalities through the Karush--Kuhn--Tucker (KKT) conditions associated with orthogonally projecting neural network outputs onto the constraint feasible region. The method is demonstrated on a continuous stirred-tank reactor (CSTR) case study for both one and two inputs. Results show that PL-KKT-hPINN preserves predictive accuracy comparable to that of a standard neural network while achieving substantially lower constraint violations. In addition, the proposed model shows improved robustness in low-data regimes, yielding lower RMSE than the unconstrained neural network for limited training sample sizes. These results demonstrate that PL-KKT-hPINN provides a computationally efficient and physically consistent framework for surrogate modeling of nonlinear chemical engineering systems.
\end{abstract}

\section{Introduction}
Surrogate modeling has been widely used as a powerful alternative to complex and computationally expensive first-principles-based mechanistic models \cite{mcbride2019overview,misener2023formulating,bhosekar2018advances,bradley2022perspectives,williams2021selection}. Different types of surrogate modeling approaches including polynomial/algebraic regression, Gaussian process regression (GP), and neural networks (NNs), have been widely used in optimization, control, and process systems engineering \cite{chen2024physics,dias2020integration,mohammadi2022surrogate}. In particular, feed-forward neural networks (FNNs), also known as multi-layer perceptrons (MLPs), have been popular because of their universal approximation capability for capturing nonlinear input-output relationships across many chemical processes \cite{bradley2022perspectives,cybenko1989approximation,hornik1989multilayer,kim2020surrogate}.
Despite their flexibility, NNs are black-box models that are not easily interpretable and may not be physically consistent with the underlying first-principles equations describing the process, such as conservation laws. This can hinder their application in decision-making problems and industrial applications \cite{chen2024physics,mukherjee2026physics}.
This becomes particularly important when surrogates are used in optimization, where the optimizer may search poorly sampled regions that require extrapolation. As a result, low training error alone does not guarantee physically feasible or truly optimal solutions for the real process. Therefore, overfitting should be avoided, and surrogate accuracy should be controlled to limit extrapolation errors \cite{misener2023formulating}.

To address this limitation, Raissi et al. \cite{raissi2019physics} proposed physics-informed neural networks (PINNs), which integrate known governing equations, such as algebraic, ordinary, or partial differential equations, into the loss function as penalty terms that penalize violations of the governing equations or constraints. This framework provides implementation flexibility by enabling the use of standard first-order learning algorithms such as stochastic gradient descent, and does not require any modification to the NN structure. However, the main limitation is that PINNs do not guarantee exact satisfaction of physical laws during training or inference. Moreover, an imbalance between the loss terms can create a challenging optimization landscape, leading to slow convergence, sensitivity to hyperparameter tuning, and limited predictive accuracy \cite{chen2024physics,mukherjee2026physics,ma2022data}.

To ensure strict satisfaction of constraints, recent research has explored hard-constrained neural networks (NNs). Beucler et al. \cite{beucler2021enforcing} introduced the equality-completion approach, where the network predicts a subset of variables known as independent variables, while the remaining variables, known as dependent variables, are analytically determined by solving a (non)linear system of equations to satisfy known algebraic constraints. This achieves exact satisfaction of selected equalities, but may require solving nonlinear equations and can introduce numerical sensitivity or additional computational overhead, depending on the constraint complexity. Amos and Kolter \cite{amos2017optnet} proposed OptNet, which embeds a differentiable optimization layer for projecting NN outputs onto the constraint feasible region by solving a quadratic program through Karush--Kuhn--Tucker (KKT) conditions within the neural network architecture. However, its reliance on iterative fixed-point solvers leads to high computational cost. Later, Agrawal et al. \cite{agrawal2019differentiable} extended this idea to convex programs and developed efficient differentiation techniques through such layers. Min and Azizan \cite{min2024hard} leveraged such optimization layers to develop HardNet-Cvx, a neural network enforcing convex constraints. A common limitation of these approaches is that the forward pass requires solving an optimization problem. Since the solution of this problem depends on the outputs of the NN, the NN must be trained simultaneously with the optimization layer, which can be computationally expensive and may limit scalability as the network and dataset size increase.

Mukherjee and Bhattacharyya \cite{mukherjee2024development} cast learning and constraint enforcement as a constrained optimization problem and train the neural network using a constrained optimization solver, such as IPOPT, instead of standard unconstrained optimization algorithms. Although these methods can handle complex constraint sets, their reliance on iterative constrained optimization imposes severe limitations for computationally demanding applications. In addition, scalability can be limited because the constrained formulation grows with the number of data points, and the solver must repeatedly solve large linear-algebra systems, leading to high computational and memory costs for large neural networks and datasets \cite{mukherjee2026physics}. Moreover, these approaches do not guarantee constraint satisfaction during inference.

To reduce computational cost relative to iterative optimization layers, Chen et al.\ \cite{chen2024physics} proposed KKT-hPINN, which enforces linear equality constraints using an orthogonal projection derived from KKT conditions. The projection is implemented through non-trainable layers that map NN predictions onto the feasible set defined by linear algebraic equations. As a result, KKT-hPINN has been shown to achieve strict constraint satisfaction and improved data efficiency, outperforming both standard NNs and soft-constrained PINNs in multiple process case studies. Since the added projection layer is non-trainable and inherently enforces constraints, KKT-hPINN is highly computationally efficient compared with the iterative approaches outlined above. The main limitation is that the framework is restricted to linear equality constraints, whereas most chemical engineering relationships, such as species balances, kinetics, and enthalpy balances, are nonlinear.

To address nonlinear algebraic constraints, Lastrucci et al.\ \cite{lastrucci2025picard} proposed Picard-KKT-hPINN, which combines local projection with a variable-freezing strategy inspired by Picard iterations to enforce nonlinear conservation laws to machine precision. 
Lastrucci et al.\ \cite{lastrucci2025enforce} further introduced Adaptive-depth Neural Projection (AdaNP), which uses an adaptive projection depth with a controllable feasibility tolerance and provable numerical stability (1-Lipschitz projection), helping prevent gradient explosion. Their approach locally linearizes the constraints and repeatedly projects onto the linearized manifold until the constraint residual falls below a prescribed tolerance 
$\epsilon$, or a maximum projection depth is reached. Iftakher et al.\ \cite{iftakher2025physics} developed KKT-HardNet, a physics-informed neural network architecture that enforces nonlinear equalities and inequalities up to machine precision. Their approach embeds a differentiable projection layer that solves the KKT system of a distance minimization problem, extending the KKT-based projection framework and using a Newton-type iterative scheme and reformulations to handle general nonlinearities. 
While these approaches strengthen constraint enforcement for nonlinear systems, a common drawback is their reliance on iterative sequential projection procedures which can increase computational cost and make them difficult to express as static non-trainable layers like KKT-hPINN.

In this study, we propose a non-iterative extension of the KKT-hPINN projection framework for nonlinear equality constraints, referred to as piecewise-linear KKT-hPINN (PL-KKT-hPINN). The key idea is to approximate the nonlinear constraint locally over multiple regions of the input domain, construct a local linear projection in each region, and then select the active regional projection using indicator functions. In this way, the method retains the efficiency and explicit structure of closed-form KKT-based projection while extending feasibility enforcement to nonlinear systems. The approach is demonstrated on a continuous stirred-tank reactor (CSTR) case study with both one-input and two-input settings.

The paper is organized as follows. Section \ref{sec:background} reviews the mathematical background of KKT-hPINNs and establishes core notation. Section \ref{sec:methods} details the proposed PL-KKT-hPINN approach. Section \ref{sec:cases} showcases PL-KKT-hPINN on two CSTR case studies. Finally, Section \ref{sec:conclusions} summarizes the main conclusions.

\section{Background and Notation}\label{sec:background}
In this section, we review core background and establish notation for NNs and KKT-hPINNs.

\subsection{Feedforward Neural Networks}
We consider NNs with \(L+1\) dense layers that each have weight matrices \(\mathbf{W}\) and bias vectors \(\mathbf{b}\). The inputs and outputs of the network are represented by $\mathbf{x}\in\mathcal{X}\subseteq \mathbb{R}^{n_x}$ and $\hat{\mathbf{y}}\in\mathbb{R}^{n_y}$, respectively.
In each layer, the output from the previous layer, \(\mathbf{z}^{(l-1)}\), is multiplied by the weight matrix \(\mathbf{W}^{(l-1)}\), shifted by the bias vector \(\mathbf{b}^{(l-1)}\), and passed through a nonlinear activation function \( \sigma \). The resulting output for that layer is \(\mathbf{z}^{(l)}\):
\begin{equation}
  \mathbf{z}^{(l)} 
  = \sigma\!\big( \mathbf{W}^{(l-1)} \mathbf{z}^{(l-1)} + \mathbf{b}^{(l-1)} \big),
  \qquad l \in \{1,\dots,L\},
\end{equation}
where notably we have that $\mathbf{z}^{(0)} = \mathbf{x}$ and $\mathbf{z}^{(L)} = \mathbf{y}$.

During training, a loss function is minimized with respect to the trainable parameters, i.e., the weights and biases \(\Theta = \{ \mathbf{W}^{l},\, \mathbf{b}^{l} \}_{l=0}^{L-1}\). In this work, the loss function is defined as the mean squared error (MSE). The standard neural network loss, $\mathcal{L}_{\mathrm{NN}}$, minimizes the average squared difference between the predicted outputs and the training data across all \(N\) training samples $\{(\mathbf{x}^{k}, \mathbf{y}^{k}) : k \in \{1,\ldots,N\}\}$:
\begin{gather}
\min_{\Theta} \; \frac{1}{2N} \sum_{k=1}^{N} \mathcal{L}_{\mathrm{NN}} ( \mathbf{x}^{k}, \mathbf{y}^{k}; \Theta) \\[8pt]
\mathcal{L}_{\mathrm{NN}} 
= \left\lVert \mathrm{NN}(\mathbf{x}^{k}; \Theta) - \mathbf{y}^{k} \right\rVert^{2} 
\label{eq:loss-nn} \\[8pt]
\hat{\mathbf{y}} = \mathrm{NN}(\mathbf{x}; \Theta) 
: \mathbb{R}^{n_x} \rightarrow \mathbb{R}^{n_y}
\end{gather}

\subsection{KKT-hPINN}
\begin{figure}[!htp]
    \centering
    \includegraphics[width=0.7\textwidth]{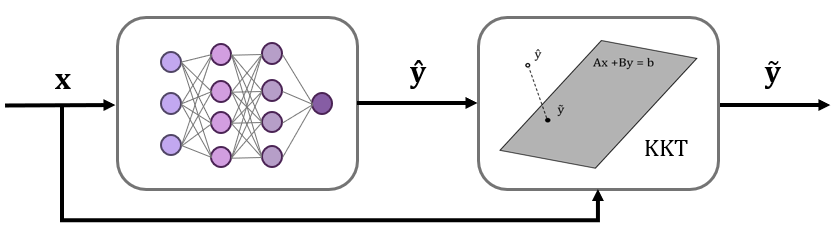}
    \caption{Schematic of the KKT-hPINN architecture for enforcing linear equality constraints. For a given input $\mathbf{x}$, the neural network first produces an unconstrained prediction $\hat{\mathbf{y}}$. This prediction is then projected onto the feasible hyperplane of the linear constraint $\mathbf{A}\mathbf{x}+\mathbf{B}\mathbf{y}=\mathbf{b}$ to obtain the corrected prediction $\tilde{\mathbf{y}}$.}
    \label{fig:KKT}
\end{figure}

The KKT-hPINN framework can be considered as a type of physics-informed neural network designed to enforce such linear constraints during both training and inference \cite{chen2024physics}. Since the NN predictions \(\hat{\mathbf{y}}\) may not inherently satisfy these constraints, the KKT-hPINN method corrects these potential violations by projecting the outputs of the network onto the linear hyperplane defined by the linear equations (i.e., the feasible region) as shown in Figure \ref{fig:KKT}. Moreover, this projection is accomplished via an non-trainable layer that is appended on to the NN.
This projection minimizes the Euclidean distance between the unconstrained predictions and the nearest point within the feasibility region for constraint satisfaction. This minimization problem can be formulated as a quadratic optimization problem subject to \(m\) linear equality constraints as follows:
\begin{equation}
\tilde{\mathbf{y}} \;=\; \arg\min_{\mathbf{y}} \; \tfrac{1}{2}\,\|\mathbf{y}-\hat{\mathbf{y}}\|^2 
\quad \text{s.t.}\quad \mathbf{A}\mathbf{x} + \mathbf{B}\mathbf{y} = \mathbf{b}
\label{eq:Projection-Quadratic}
\end{equation}
where $\mathbf{A} \in \mathbb{R}^{m \times n_x}$, $\mathbf{B} \in \mathbb{R}^{m \times n_y}$, and $\mathbf{b} \in \mathbb{R}^{m}$ are constants that define the equations. In chemical engineering, these equations commonly capture linear conservation laws such as total mass balances.

The optimal primal solution $\tilde{\mathbf{y}}$ and the corresponding dual solution given by the Lagrange multipliers $\lambda^{*}$ for Problem \eqref{eq:Projection-Quadratic} satisfy the first-order KKT conditions:
\begin{equation}
\begin{bmatrix}
\mathbf{I} & \mathbf{B}^{\!\top} \\
\mathbf{B} & \mathbf{0}
\end{bmatrix}
\begin{bmatrix}
\tilde{\mathbf{y}} \\
\lambda^{*}
\end{bmatrix}
=
\begin{bmatrix}
\hat{\mathbf{y}} \\
\mathbf{b} - \mathbf{A}\mathbf{x}
\end{bmatrix}
\label{eq:KKT}
\end{equation}
which are rearranged to obtain a closed-form analytic expression for $\tilde{\mathbf{y}}$:
\begin{equation}
\tilde{\mathbf{y}} = \mathbf{A}^{*}\mathbf{x} + \mathbf{B}^{*}\hat{\mathbf{y}} + \mathbf{b}^{*}
\label{eq:kkt_projection_solution}
\end{equation}
where \(\mathbf{A}^{*}\), \(\mathbf{B}^{*}\) and \(\mathbf{b}^{*}\) are fixed constant matrices and vectors defined by:
\begin{gather}
\mathbf{A}^{*} = -\,\mathbf{B}^{\!\top}(\mathbf{B}\mathbf{B}^{\!\top})^{-1}\mathbf{A}
\label{eq:kkt_Astar} \\
\mathbf{B}^{*} = \mathbf{I} - \mathbf{B}^{\!\top}(\mathbf{B}\mathbf{B}^{\!\top})^{-1}\mathbf{B}
\label{eq:kkt_Bstar} \\
\mathbf{b}^{*} = \mathbf{B}^{\!\top}(\mathbf{B}\mathbf{B}^{\!\top})^{-1}\mathbf{b}.
\label{eq:kkt_bstar}
\end{gather}

The resulting projection matrices are used in two non-trainable layers with fixed parameters \(\mathbf{A}^{*}\), \(\mathbf{B}^{*}\), and \(\mathbf{b}^{*}\) to the neural network which respective outputs are added together. Thus, the addition of the two layers to the network serves to correct the potential violation of constraints by the neural network's predictions, \(\mathrm{NN}( \mathbf{x}^{k}; \Theta)\), by projecting them onto the feasible
region defined by the linear equality constraints. Note, that this approach is general such that these projection layers can be added to any machine learning model, not just NNs.

Following the addition of the two layers, the predictions of the neural network are not fed directly into the loss function, but they are instead first projected onto the feasible region defined by the linear constraints. Then, the outputs of the projection layer \(\tilde{\mathbf{y}} = \mathbf{A}^{*}\mathbf{x} + \mathbf{B}^{*}\hat{\mathbf{y}} + \mathbf{b}^{*}\), are passed to the loss function \(\mathcal{L}_{\mathrm{KKT}}\) which is defined:
\begin{equation}
\mathcal{L}_{\mathrm{KKT}}= \left\lVert \mathbf{A}^{*}\mathbf{x}^{k}+ \mathbf{B}^{*}\,\mathrm{NN}(\mathbf{x}^{k}; \Theta)+ \mathbf{b}^{*}- \mathbf{y}^{k}\right\rVert^{2}.
\label{eq:loss-kkt}
\end{equation}
Hence, the addition of the projection layer changes the learning direction represented by the gradient of the loss function with respect to the network parameters, yielding different NN parameters $\Theta$:
\begin{equation}
\min_{\Theta} \; \frac{1}{2N} \sum_{k=1}^{N} \mathcal{L}_\mathrm{KKT}( \mathbf{x}^{k}, \mathbf{y}^{k}; \Theta).
\end{equation}

\section{PL-KKT-hPINN Framework} \label{sec:methods}
In this section, we detail the PL-KKT-hPINN framework and discuss practical considerations for its implementation. 

\subsection{Architectural Overview} \label{subsec:overall_approach}
Consider an unconstrained NN \(\hat{\mathbf{y}} = \mathrm{NN}(\mathbf{x};\Theta) \in \mathbb{R}^{n_y}\) for which we seek to minimally adjust the outputs $\hat{\mathbf{y}}$ such that they satisfy a set of \(m\) nonlinear equality constraints:
\begin{equation}
g_i(\mathbf{x},\mathbf{y})=0, \qquad i\in\{1,\ldots,m\}
\label{eq:nonlinear_constraints}
\end{equation}
which we express more compactly as:
\begin{equation}
\mathbf{g}(\mathbf{x},\mathbf{y})=\mathbf{0}
\label{eq:nonlinear_constraints_vector}
\end{equation}
with the vector-valued function $\mathbf{g}:\mathbb{R}^{n_x}\times\mathbb{R}^{n_y}\rightarrow\mathbb{R}^{m}$. This minimal adjustment of \(\hat{\mathbf{y}}\) is defined by the solution to the nonlinear projection problem:
\begin{equation}
\begin{aligned}
\tilde{\mathbf{y}} = &&\amin_{\mathbf{y}} &&& \frac{1}{2}\left\lVert \mathbf{y}-\hat{\mathbf{y}} \right\rVert^2\\
&&\text{s.t.} &&&\mathbf{g}(\mathbf{x},\mathbf{y})=\mathbf{0}.
\end{aligned}
\label{eq:nonlinear_projection}
\end{equation}
This formulation generalizes Problem~\eqref{eq:Projection-Quadratic} to use nonlinear equality constraints; however, a closed-form solution cannot likewise be obtained via its KKT conditions for general nonlinear equality constraints. 

Hence, for PL-KKT-hPINN, we approximate the nonlinear constraints $\mathbf{g}(\cdot)=0$ by a set of piecewise-linear equality constraints which individually exhibit closed-forms for orthogonal projection following Equation \eqref{eq:kkt_projection_solution}. This piecewise-linear approximation is comprised of \(p\) groups of linear equality constraints:
\begin{equation}
\mathbf{A}_j\mathbf{x}+\mathbf{B}_j\mathbf{y}=\mathbf{b}_j,\qquad j\in\mathcal{J}=\{1,\ldots,p\}
\label{eq:regional_linear_constraint}
\end{equation}
that are comprised of constants $\mathbf{A}_j \in \mathbb{R}^{m\times n_x}$, $\mathbf{B}_j \in \mathbb{R}^{m\times n_y}$, and $\mathbf{b}_j \in \mathbb{R}^{m}$ and are each enforced over a subregion $R_j \subset \mathcal{X}$. The subregions are non-overlapping such that:
\begin{equation}
\bigcup_{j\in\mathcal{J}} R_j = \mathcal{X}
\label{eq:domain_partition_general}
\end{equation}
and
\begin{equation}
R_j \cap R_{j'} = \emptyset, \quad (j, j') \in \mathcal{J} \times \mathcal{J}, \quad j\neq j'.
\label{eq:nonoverlapping}
\end{equation}
Applying this piecewise-linear representation to Problem \eqref{eq:nonlinear_projection}, we obtain the approximate reformulation (expressed using generalized disjunctive programming to describe the piecewise-linear constraints \cite{ammari2023linear}):
\begin{equation}
\begin{aligned}
\tilde{\mathbf{y}} = &&\amin_{\mathbf{y}, Q} &&&\frac{1}{2}\left\lVert \mathbf{y}-\hat{\mathbf{y}} \right\rVert^2 \\
&&\text{s.t.} &&& \bigvee_{j \in \mathcal{J}}\begin{bmatrix} Q_j \\ \mathbf{A}_j\mathbf{x}+\mathbf{B}_j\mathbf{y}=\mathbf{b}_j \\
\mathbf{x}\in R_j
\end{bmatrix}\\
&&&&& \text{exactly}(1, Q) = \text{True}
\end{aligned} 
\label{eq:gdp_projection_problem}
\end{equation}
where $Q \in \{\text{True}, \text{False}\}$ are logical variables that indicate which subregion $R_j$ of linear constraints is active. Following standard properties of piecewise-approximations, we expect this approximation to become increasing exact with increased $p$, though the convergence rate is dependent on the method used to derive the piecewise-linear representation. More discussion on defining this representation and choosing $p$ is provided in Section \ref{subsec:piecewise_linear_approximation}.

Problem \eqref{eq:gdp_projection_problem} can be recast as $p$ subproblems that correspond to each disjunct being potentially enforced:
\begin{equation}
\begin{aligned}
\tilde{\mathbf{y}}_j = &&\amin_{\mathbf{y}} &&&\frac{1}{2}\left\lVert \mathbf{y}-\hat{\mathbf{y}} \right\rVert^2 \\
&&\text{s.t.} &&& \mathbf{A}_{j}\mathbf{x}+\mathbf{B}_{j}\mathbf{y}=\mathbf{b}_{j}
\end{aligned} 
\label{eq:gdp_subproblems}
\end{equation}
which is identical in form to Problem \eqref{eq:Projection-Quadratic} and likewise exhibits the closed-form solution:
\begin{equation}
\tilde{\mathbf{y}}_{j} = \mathbf{A}^{*}_{j}\mathbf{x} + \mathbf{B}^{*}_{j}\hat{\mathbf{y}} + \mathbf{b}^{*}_{j}
\label{eq:regional_projection_solution}
\end{equation}
where \(\mathbf{A}^{*}_{j}\), \(\mathbf{B}^{*}_{j}\), and \(\mathbf{b}^{*}_{j}\) are the fixed projection parameters obtained by applying Equations~\eqref{eq:kkt_Astar}--\eqref{eq:kkt_bstar} to the local linear constraints in region \(R_j\). Since $\mathbf{x}$ is a parametric input to Problem \eqref{eq:gdp_projection_problem}, we have that $Q_j = \text{True} \iff \mathbf{x} \in R_j$ which means that the solution of Problem \eqref{eq:gdp_projection_problem} is given by Equation \eqref{eq:regional_projection_solution} for the index $j$ that corresponds to the region $R_j$ that contains $\mathbf{x}$. Thus, using an indicator function $\Omega_{R_j} : \mathcal{X} \mapsto \{0, 1\}$ which indicates whether $\mathbf{x} \in R_j$, we can equivalently reformulate Problem \eqref{eq:gdp_projection_problem} as:
\begin{equation}
\tilde{\mathbf{y}} = \sum_{j\in\mathcal{J}} \Omega_{R_j}(\mathbf{x}) \tilde{\mathbf{y}}_{j}
\label{eq:pl_kkt_sum}
\end{equation}
since by Equation \eqref{eq:nonoverlapping} we have that the regions $R_j$ are non-overlapping. The selection and setup of an appropriate indicator function is discussed in Section \ref{subsec:indicator_functions}.

\begin{figure}[!htb]
    \centering
    \includegraphics[width=0.95\textwidth]{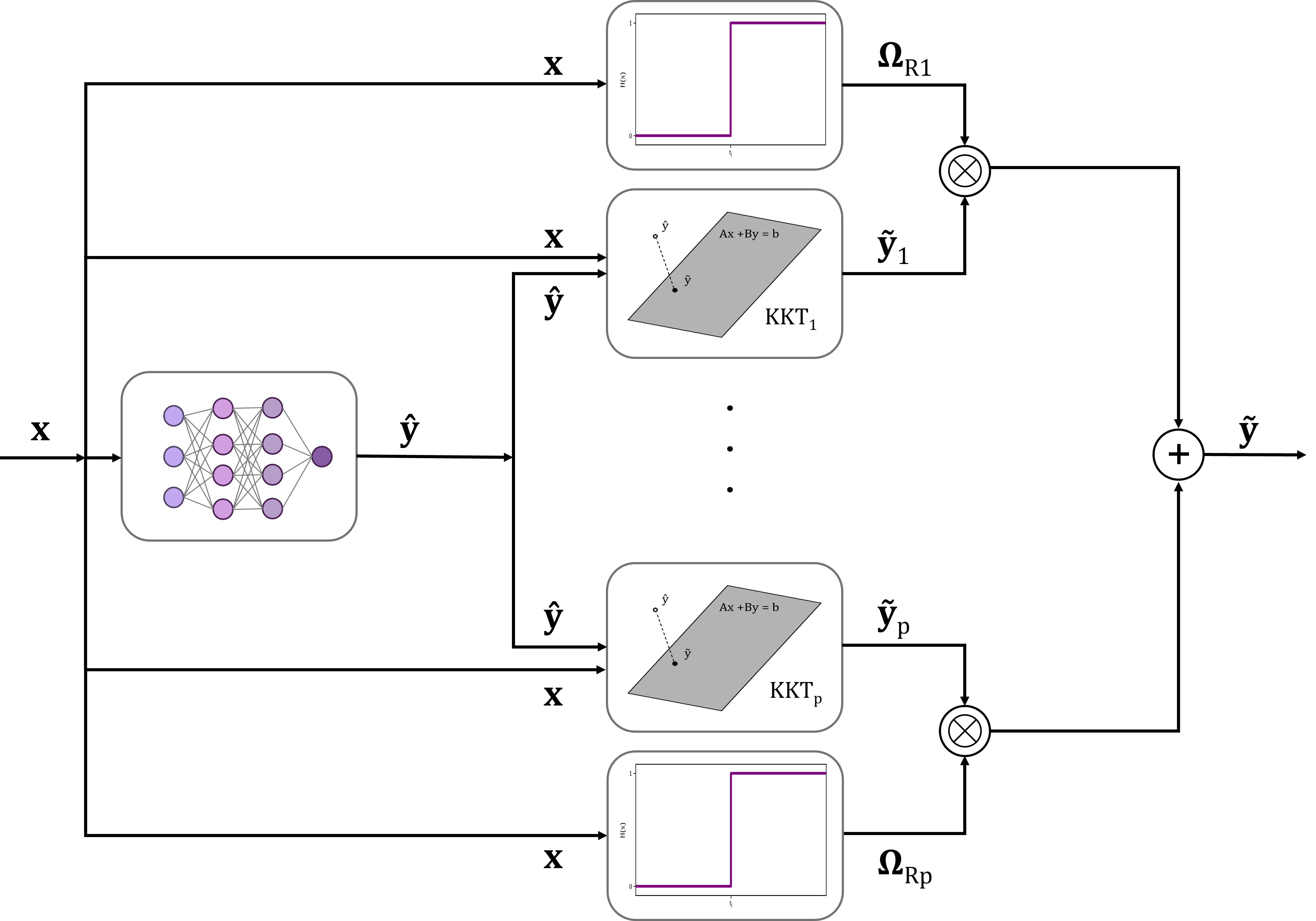}
    \caption{Schematic of the proposed PL-KKT-hPINN architecture for enforcing nonlinear equality constraints using piecewise-linear KKT projections. For a given input $\mathbf{x}$, the neural network first produces an unconstrained prediction $\hat{\mathbf{y}}$. In each linearization region $R_j$, this prediction is projected onto the local feasible hyperplane defined by the linearized constraint $\mathbf{A}_j\mathbf{x}+\mathbf{B}_j\mathbf{y}=\mathbf{b}_j$ to obtain the regional corrected prediction $\tilde{\mathbf{y}}_{j}$. The regional predictions are then combined using an indicator functions $\Omega_{R_j}(\mathbf{x})$ to obtain the final constrained prediction $\tilde{\mathbf{y}}$.}
    \label{fig:PL-KKThPINN}
\end{figure}

With Equation \eqref{eq:pl_kkt_sum}, we can adapt the KKT-hPINN architecture shown in Figure \ref{fig:KKT} to be applied in parallel across all $p$ projections and then aggregate all the $\tilde{\mathbf{y}}_j$ signals using an appropriate indicator function $\Omega_{R_j}$ and sum to obtain $\tilde{\mathbf{y}}$. This architecture for PL-KKT-hPINN is depicted in Figure \ref{fig:PL-KKThPINN}. Unlike other approaches in the literature (e.g., AdaNP) which project nonlinear equalities via a dynamic number of sequential linear projections, PL-KKT-hPINN projects using a fixed number of parallel linear projections. This parallel structure enables efficient vectorized computation and maintains a static NN architecture. Moreover, like other projection-based approaches PL-KKT-hPINN, is agnostic of whether a neural network is used and in principle can be added as a projection layer to any machine learning model.

The trainable parameters are the weights and biases of the inner neural network, while the projection parameters 
\(\mathbf{A}^{*}_{j}\), \(\mathbf{B}^{*}_{j}\), and \(\mathbf{b}^{*}_{j}\) are fixed and non-trainable. The model is trained by minimizing the mean squared error between the projected prediction $\tilde{\mathbf{y}}$ and the training data:
\begin{equation}
\min_{\Theta}\; \frac{1}{2N} \sum_{k=1}^{N} \mathcal{L}_{\mathrm{PL\text{-}KKT}}(\mathbf{x}^{k},\mathbf{y}^{k};\Theta)
\end{equation}
where we define the loss function $\mathcal{L}_{\mathrm{PL\text{-}KKT}}$ as:
\begin{equation}
\mathcal{L}_{\mathrm{PL\text{-}KKT}} =
\left\lVert
\sum_{j=1}^{p}
\Omega_{R_j}(\mathbf{x}^{k})
\left(
\mathbf{A}^{*}_{j}\mathbf{x}^{k}
+
\mathbf{B}^{*}_{j}\mathrm{NN}(\mathbf{x}^{k};\Theta)
+
\mathbf{b}^{*}_{j}
\right)
-
\mathbf{y}^{k}
\right\rVert^{2}.
\label{eq:loss_pl_kkt}
\end{equation}
Note that this loss function does not require that the indicator function $\Omega_{R_j}$ be differentiable since $\Omega_{R_j}$ only depends on $\mathbf{x}^k$ which are fixed during training.

This approach also can help to improve data efficiency for training, since the projection reduces the admissible output space to one that is locally consistent with the imposed constraints. In this sense, the constraint projection acts as a structural regularization mechanism to helps to reduce overfitting. Therefore, there advantages of projection-based methods like PL-KKT-hPINN are often most evident in low-data regimes where an unconstrained neural network is more likely to overfit. the available data or produce physically inconsistent predictions. However, projection-based methods like PL-KKT-hPINN exhibit significantly increased constraint satisfaction across all dataset sizing. These behaviors are demonstrated in Section \ref{sec:cases}.

\subsection{Determining the Piecewise-Linear Approximation} \label{subsec:piecewise_linear_approximation}

Several approaches have been proposed in the literature to obtain piecewise-linear approximations of multi-dimensional nonlinear functions such as linearizations based on Taylor-series and representations derived from linear decision trees \cite{misener2010piecewise, ammari2023linear}. In principle, any piecewise-linear approach can be used to model nonlinear equality constraints with PL-KKT-hPINN so long as it can be expressed using Equations \eqref{eq:regional_linear_constraint}-\eqref{eq:nonoverlapping}. In this work, we limit our scope to using first-order Taylor-series approximation to construct local linear constraints in the form of Equation \eqref{eq:regional_linear_constraint} defined over partitioned input subregions $R_j$.

For a one-dimensional input \(\mathbf{x}\in\mathbb{R}^{1}\), the input domain \(\mathcal{X}\) is divided into \(p\) intervals,
\begin{equation}
R_j=[e_{j-1},e_j],
\qquad j \in \mathcal{J}
\label{eq:one_dim_regions}
\end{equation}
where \(\{e_0,e_1,\ldots,e_p\}\) are the region boundaries. The boundaries can be chosen at uniform intervals for simplicity or nonuniform intervals to better capture areas of increased nonlinearity in $g(\mathbf{x}, \mathbf{y})$. For a multi-dimensional input $\mathbf{x} \in \mathbb{R}^{n_x}$, the input domain can be represented via the Cartesian product of one-dimensional intervals to form hyperrectangular regions:
\begin{equation}
R_j=[e_{q,j-1},e_{q,j}]^{n_x},\qquad j \in \mathcal{J}, q \in \{1, \dots, n_x\}
\label{eq:multi_dim_regions}
\end{equation}
where $q$ indexes each of $n_x$ dimensions of $\mathbf{x}$.
The number of linearization regions controls the trade-off between approximation accuracy and computational cost. With too few regions, the local linear approximation may be too coarse to represent the nonlinear constraint over the full input domain, leading to larger nonlinear constraint violations. Naturally, the piecewise-linear approximation error (the only approximation error introduced by the projection in PL-KKT-hPINN) tends to zero as the number of regions increases. However, using more regions also linearly increases the computation cost of PL-KKT-hPINN as shown in Section \ref{sec:cases}.

Next, we select a representative center point $(\mathbf{x}_j^c,\mathbf{y}_j^c)$ for each region $R_j$ from the set of training data points $\{(\mathbf{x}^k,\mathbf{y}^k) : k \in \{1,\dots,N\}\}$. A common choice for $\mathbf{x}_j^c$ is for it to be at (or near) the geometric midpoint of $R_j$ (i.e., $\mathbf{x}_j^c = \frac{1}{2}(e_j - e_{j-1})$ if $n_x = 1$). Note that we restrict our selection to a training data point to avoid needing to approximate $\mathbf{y}_j^c$ which would introduce another source of approximation error to the piecewise-linear representation. 
Centered on  $(\mathbf{x}_j^c, \mathbf{y}_j^c)$, each nonlinear equality constraint $g_i(\mathbf{x},\mathbf{y})=0$ is then approximated via a first-order Taylor expansion such that for $R_j$ the function is approximated as:
\begin{equation}
g_i(\mathbf{x},\mathbf{y}) \approx g_i\!\left(\mathbf{x}_j^c,\mathbf{y}_j^c\right) + \nabla_\mathbf{x} g_i(\mathbf{x}_j^c, \mathbf{y}_j^c)^T \left(\mathbf{x}-\mathbf{x}_j^c\right) + \nabla_\mathbf{y} g_i(\mathbf{x}_j^c, \mathbf{y}_j^c)^T \left(\mathbf{y}-\mathbf{y}_j^c\right),
\label{eq:taylor_scalar_constraint}
\end{equation}
where $\nabla_\mathbf{x} g_i: \mathcal{X} \times \mathbb{R}^{n_y} \mapsto \mathbb{R}^{n_x}$ and $\nabla_\mathbf{y} g_i: \mathcal{X} \times \mathbb{R}^{n_y} \mapsto \mathbb{R}^{n_y}$ are the gradients of $g_i$ with respect to $\mathbf{x}$ and $\mathbf{y}$, respectively. We use these gradients to define the local approximation of all $m$ nonlinear constraints in $R_j$ in the form of Equation \eqref{eq:regional_linear_constraint} by defining:
\begin{equation}
\mathbf{A}_j =
\begin{bmatrix}
\nabla_\mathbf{x} g_1(\mathbf{x}_j^c, \mathbf{y}_j^c)^T\\
\nabla_\mathbf{x} g_2(\mathbf{x}_j^c, \mathbf{y}_j^c)^T\\
\vdots\\
\nabla_\mathbf{x} g_m(\mathbf{x}_j^c, \mathbf{y}_j^c)^T
\end{bmatrix},
\quad
\mathbf{B}_j
=
\begin{bmatrix}
\nabla_\mathbf{y} g_1(\mathbf{x}_j^c, \mathbf{y}_j^c)^T\\
\nabla_\mathbf{y} g_2(\mathbf{x}_j^c, \mathbf{y}_j^c)^T\\
\vdots\\
\nabla_\mathbf{y} g_m(\mathbf{x}_j^c, \mathbf{y}_j^c)^T
\end{bmatrix}, \quad
\mathbf{b}_j = \mathbf{A}_j\mathbf{x}_j^c + \mathbf{B}_j\mathbf{y}_j^c -
\begin{bmatrix}
g_1\!\left(\mathbf{x}_j^c,\mathbf{y}_j^c\right)\\
g_2\!\left(\mathbf{x}_j^c,\mathbf{y}_j^c\right)\\
\vdots\\
g_m\!\left(\mathbf{x}_j^c,\mathbf{y}_j^c\right)
\end{bmatrix}.
\label{eq:regional_A_B_def}
\end{equation}
The accuracy of this approximation can be estimated via evaluation of the truncated higher order Taylor-expansion terms and/or sample average approximation of the constraint violation using the training dataset. Also, any constraint $g_i(\cdot) = 0$ that is affine is exactly represented as demonstrated in Section \ref{sec:cases}. One caveat to this approach is that large gradient evaluations can induce significant sensitivity in constraint satisfaction relative to small changes in the corresponding input \cite{ralph1997sensitivity}. This can be alleviated by reformulation and/or adjusting the location region boundaries. 

\subsection{Indicator Function Definition}
\label{subsec:indicator_functions}

To implement Equation \eqref{eq:pl_kkt_sum} in PL-KKT-hPINN, we need to define $\Omega_{R_j}(\mathbf{x})$ such that it can be be readily embedded in standard deep learning modeling languages such as PyTorch. Again, we note that $\Omega_{R_j}(\mathbf{x})$ does not need to be differentiable to facilitate training the PL-KKT-hPINN model since it depends only on the input \(x\), which is fixed during training, and not on the trainable NN parameters. This means that the gradient of the loss function with respect to the NN parameters is multiplied by the activation functions. Thus, we use nonsmooth step functions to exactly express $\Omega_{R_j}(\mathbf{x})$ and avoid introducing additional approximation error beyond the piecewise-linear representation of the nonlinear constraints.

For a one-dimensional input domain divided into \(p\) non-overlapping regions according to Equation \eqref{eq:one_dim_regions}, the indicator can be defined using a Heaviside function $H : \mathbb{R} \mapsto \{0, 1\}$:
\begin{equation}
\Omega_{R_j}(\mathbf{x})=
\begin{cases}
1, & x\in R_j,\\
0, & x\notin R_j.
\end{cases}  = H(\mathbf{x}-e_{j-1})-H(\mathbf{x}-e_j)
\label{eq:heaviside}
\end{equation}
where $H$ is defined with a scalar input $z \in \mathbb{R}$:
\begin{equation}
H(z) = \begin{cases}
1, & z \geq 0 \\
0, & z < 0
\end{cases}.
\end{equation}
This definition of Heaviside ensures that each input $\mathbf{x} \in \mathcal{X}$ corresponds to exactly one linearization region. For multi-dimensional regions comprised of independent one-dimensional regions as defined in Equation \eqref{eq:multi_dim_regions}, we can express the activation function as a product of Heaviside functions over each dimension of $\mathbf{x}$ since the region is defined by a Cartesian product:
\begin{equation}
\Omega_{R_j} = \prod_{q = 1}^{n_x} \bigg(H(x_q-e_{q,j-1})-H(x_q-e_{q,j})\bigg)
\label{eq:multi_indicator}
\end{equation}
which produces Equation \eqref{eq:heaviside} in the special case that $n_x = 1$. Other types of multi-dimensional regions are also compatible with the PL-KKT-hPINN, but we leave the study of appropriate activation function representations for such regions to future work.

\section{Case Studies} \label{sec:cases}
In this section, we illustrate the proposed PL-KKT-hPINN framework via two steady-state CSTR case studies. All experiments were implemented in Python 3.12.3 using PyTorch 2.7.0 and trained with the Adam optimizer on CPU. The computations were performed on a Linux server running Ubuntu 24.04.3 LTS, equipped with an Intel Xeon W5-3435X processor with 16 physical cores, 32 logical threads, and 503 GiB of RAM. Each trained model instance is replicated 50 times with different random initializations to account for the stochasticity during training. The source code used to generate the results is available at \url{https://github.com/PULSI-Opt/PL-KKT-hPINN}.

\subsection{Case Study Definition and Configuration}
We consider a steady-state CSTR with the reversible reaction:
\begin{equation}
\ce{A + 2B <=> C}
\end{equation}
with a rate law for species A expressed by:
\begin{equation}
r_A = - \; k_f \, C_A \, C_B^2 \;+\; k_r \, C_C \label{eq:rA}
\end{equation}
where $C_A$, $C_B$, and $C_C$ are the molar concentration of species A, B, and C, respectively. The rate constants $k_f$ and $k_r$ at the reactor temperature $T$ are described using the Arrhenius equations:
\begin{align}
k_f(T) &= A_{f0} \, \exp\!\left(-\frac{E_{af}}{R T}\right) \label{eq:kf}\\
k_r(T) &= A_{r0} \, \exp\!\left(-\frac{E_{ar}}{R T}\right) \label{eq:kr}
\end{align}
where $A_{f0}$ is the forward pre-exponential factor, $A_{r0}$ is the reverse pre-exponential factor, $E_{af}$ is the forward activation energy, $E_{ar}$ is the reverse activation energy, and $R$ is the ideal gas constant. We then have that the steady-state mole balance for species A in the CSTR is:
\begin{equation}
C_{A0} - C_A + r_A\tau = 0
\label{eq:A_mole_balance}
\end{equation}
where $\tau$ is the residence time and $C_{A0}$ is the feed concentration of A. Similarly, we can derive the mole balance for species B as:
\begin{equation}
C_{B0} - C_B + 2r_A\tau = 0.
\label{eq:B_mole_balance}
\end{equation}
Assuming uniform density, the total mass balance is expressed:
\begin{equation}
C_{A0}+C_{B0}+C_{C0} \;=\; C_A + C_B + C_C.
\label{eq:total_mass}
\end{equation}

\begin{table}[h!]
\centering
\caption{Physical and Kinetic Parameters of the CSTR Unit}
\label{tab:cstr_constants}
\begin{tabular}{l l c c}
\hline
\textbf{Symbol} & \textbf{Description} & \textbf{Value} & \textbf{Unit} \\
\hline
$C_{B0}$ & Feed concentration of B &  2.0 & mol/L \\
$C_{C0}$ & Feed concentration of C &  0.0 & mol/L \\
$\tau$ & Residence time & 10.0 & s \\
$A_{f0}$ & Forward Arrhenius pre-exponential factor & $1\times10^{13}$ & s$^{-1}$ \\
$E_{af}$ & Forward activation energy & 90000 & J/mol \\
$A_{r0}$ & Reverse Arrhenius pre-exponential factor & $1\times10^{11}$ & s$^{-1}$ \\
$E_{ar}$ & Reverse activation energy & 80000 & J/mol \\
$R$ & Universal gas constant & 8.314 & J/(mol·K) \\
\hline
\end{tabular}
\end{table}

In silico training data is generated by solving Equations \eqref{eq:A_mole_balance}-\eqref{eq:total_mass} for $C_A$, $C_B$, and $C_C$ over a range of $C_{A0}$ and $T$ values using the constant values reported in Table \ref{tab:cstr_constants}. In particular, two cases are considered: a 1D case where $C_{A0}$ is varied on the interval $[0.5M, 1.5M]$ with $T = 350K$ and a 2D case where $C_{A0} \in [0.8M, 1.2M]$ with $T \in[280K, 460K]$. Thus, we have outputs $\mathbf{y} = [C_A \; C_B \; C_C]^T$ with inputs $\mathbf{x} = [C_{A0}]$ and $\mathbf{x} = [C_{A0} \;T]^T$ for the two cases, respectively. To test the ability of PL-KKT-hPINN to enforce a system of nonlinear and linear equality constraints, we choose to enforce the molar balance described in Equation \eqref{eq:A_mole_balance} and the total mass balance from Equation \eqref{eq:total_mass}:
\begin{equation}
\begin{gathered}
g_1(T,C_{A0},C_A,C_B, C_C) = C_{A0} - C_A - k_f(T) \cdot C_A C_B^{2} \tau + k_r(T) \cdot C_C \tau =0 \\
g_2(C_{A0},C_A,C_B, C_C) = C_{AO} - C_A + C_{B0} - C_B + C_{C0} - C_C = 0.
\end{gathered}
\label{eq:f_nonlin}
\end{equation}
Equation \eqref{eq:B_mole_balance} is not enforced as a constraint such that the model is not fully specified by the constraints. For PL-KKT-hPINN, Equation \eqref{eq:f_nonlin} is approximated using the Taylor expansion described in Equation \eqref{eq:taylor_scalar_constraint} over uniformly spaced intervals defined on the domains of $C_{A0}$ and $T$.

The same NN model is used for the standard NNs, PL-KKT-hPINNs, and PINNs considered in this case study. Namely the NN models all use two hidden dense layers that each use 32 neurons with rectified linear unit activation functions. These are all trained using a constant learning rate of \(10^{-4}\) over 1000 epochs. No additional hyperparameter tuning was performed
because the training and validation curves consistently showed stable convergence for the CSTR case study. In each configuration, the dataset is divided into training, validation, and testing subsets using a 60/20/20 split. Unless otherwise stated, 150 samples and 30 linearization regions are used for the 1D case, and 170 samples and 21 linearization regions are used for the 2D case.

\subsection{Model Training Performance}

\begin{figure}[!htb]
    \centering

    \begin{subfigure}[t]{0.48\textwidth}
        \centering
        \includegraphics[width=\linewidth]{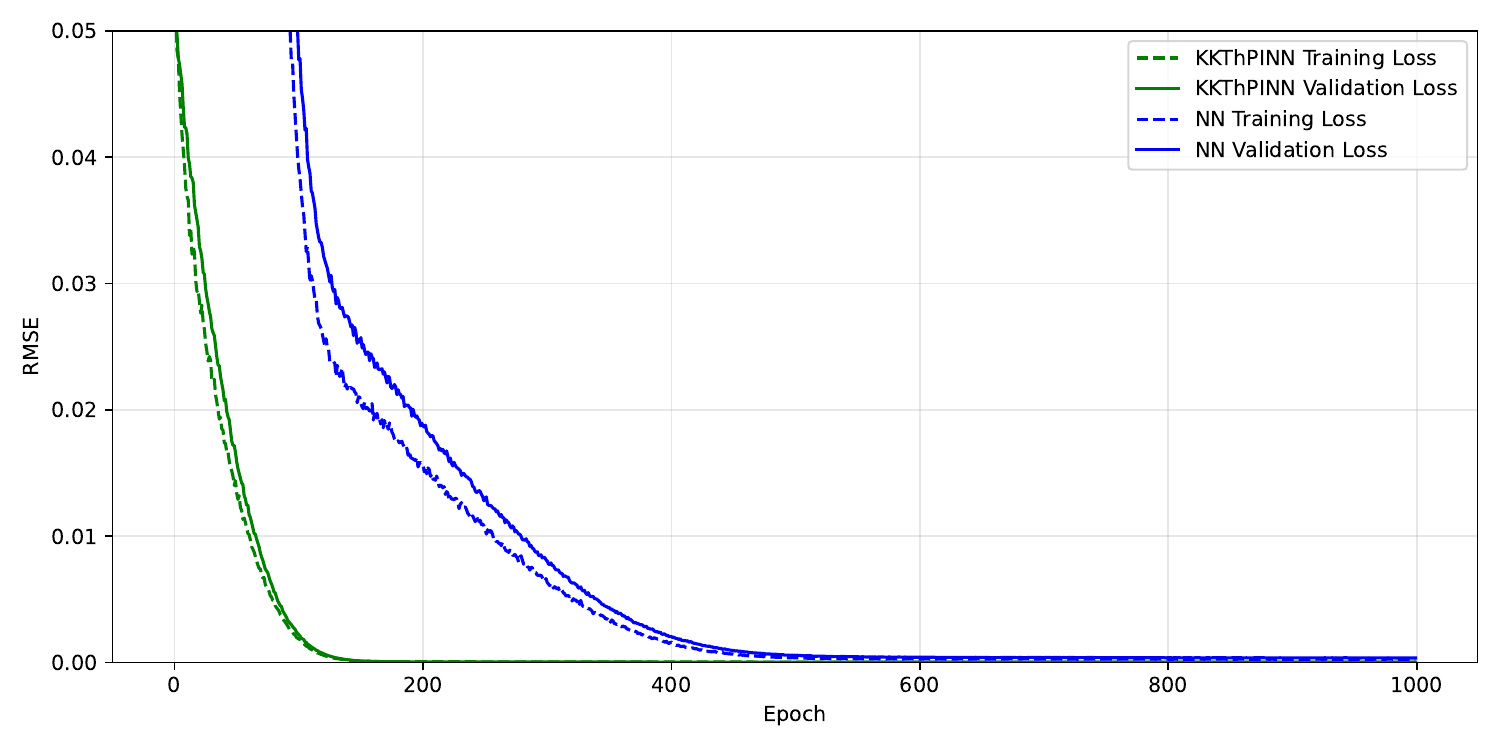}
        \caption{Training and validation RMSE}
        \label{fig:loss-PL}
    \end{subfigure}
    \hfill
    \begin{subfigure}[t]{0.48\textwidth}
        \centering
        \includegraphics[width=\linewidth]{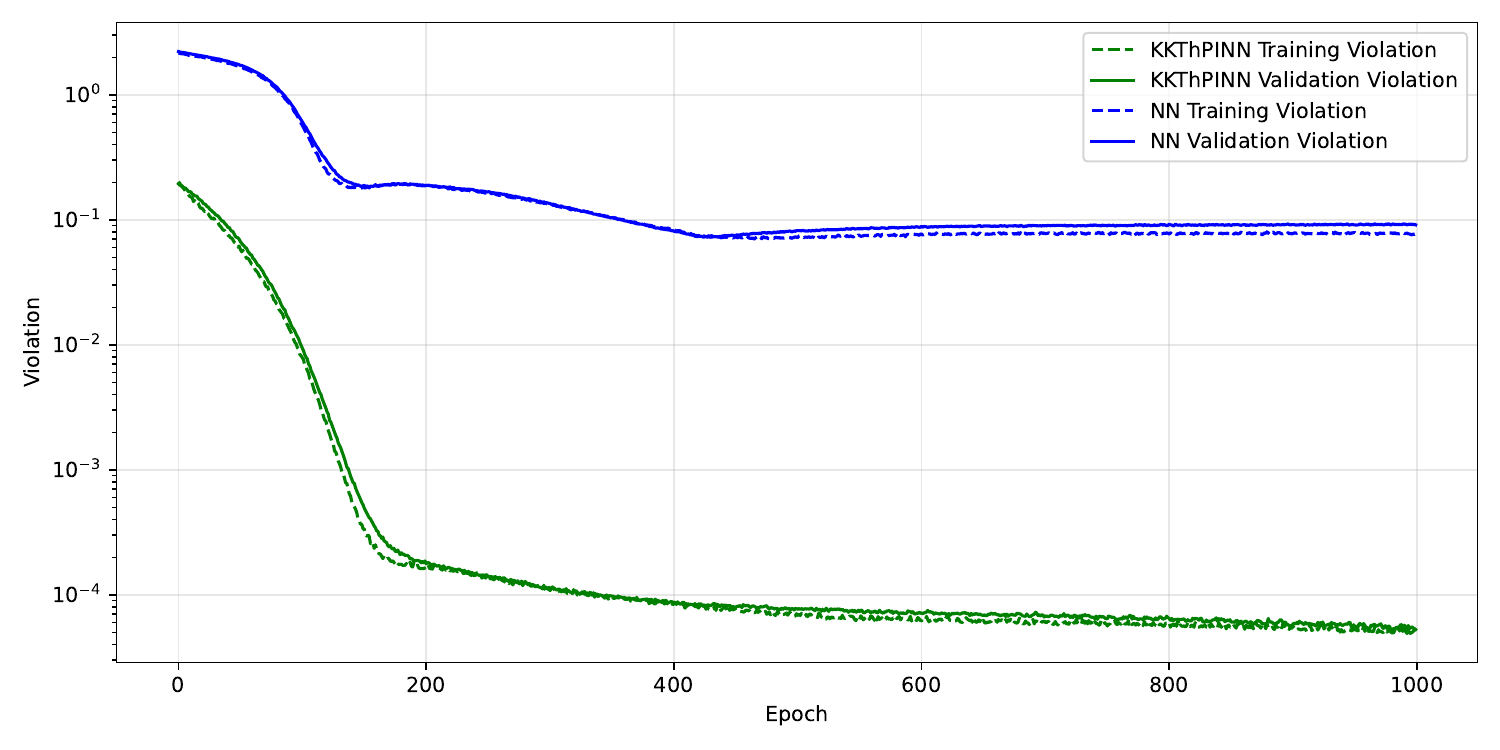}
        \caption{Nonlinear constraint violation}
        \label{fig:violation-PL}
    \end{subfigure}

    \caption{Training behavior of the standard neural network (NN) and the proposed PL-KKT-hPINN for the 1D CSTR case study. The panels show: (a) training and validation RMSE over epochs, and (b) training and validation nonlinear constraint violation over epochs.}
    \label{fig:training_behavior_1D}
\end{figure}

\begin{figure}[!htb]
    \centering

    \begin{subfigure}[t]{0.48\textwidth}
        \centering
        \includegraphics[width=\linewidth]{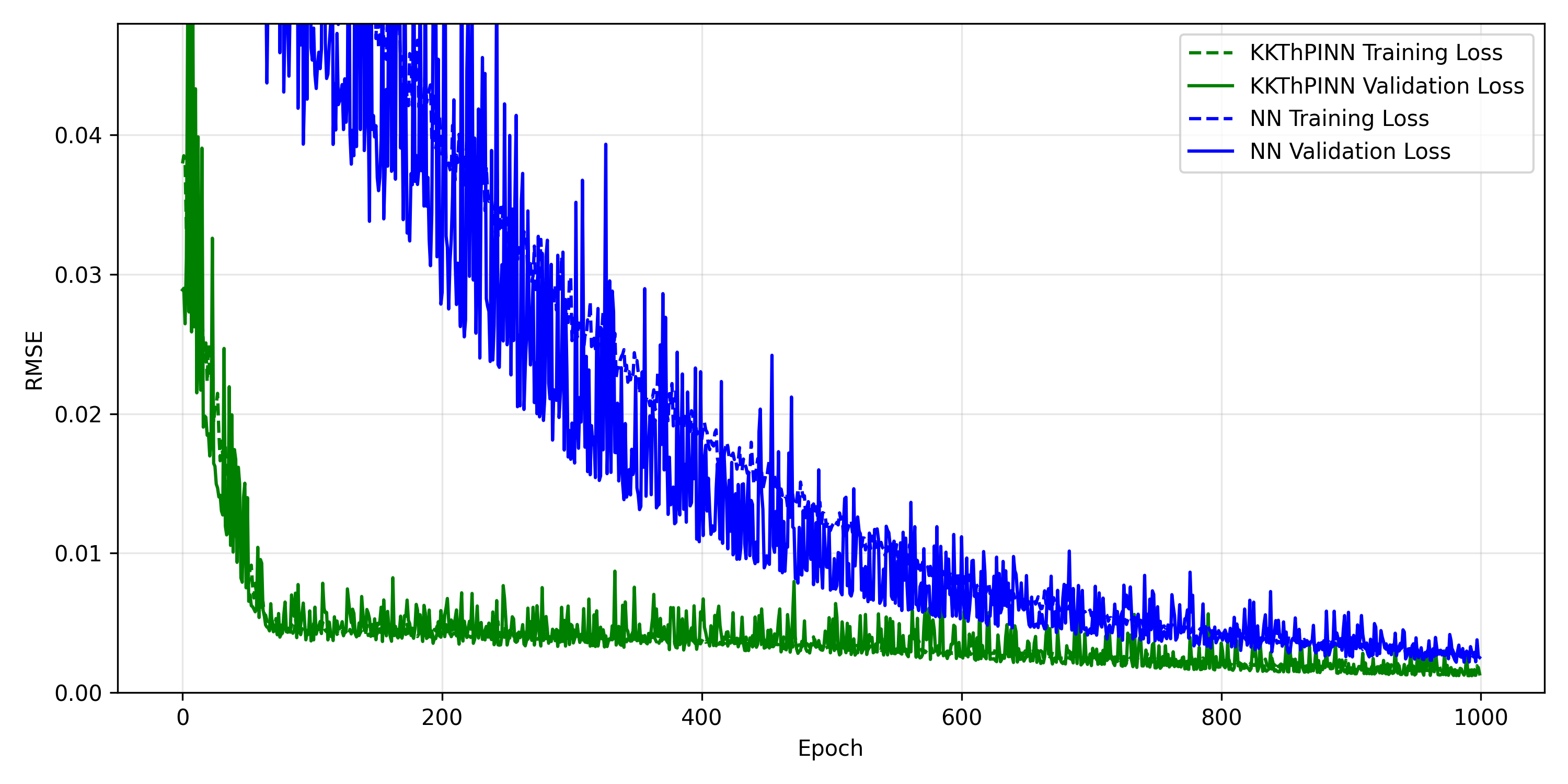}
        \caption{Training and validation RMSE}
        \label{fig:loss-2D}
    \end{subfigure}
    \hfill
    \begin{subfigure}[t]{0.48\textwidth}
        \centering
        \includegraphics[width=\linewidth]{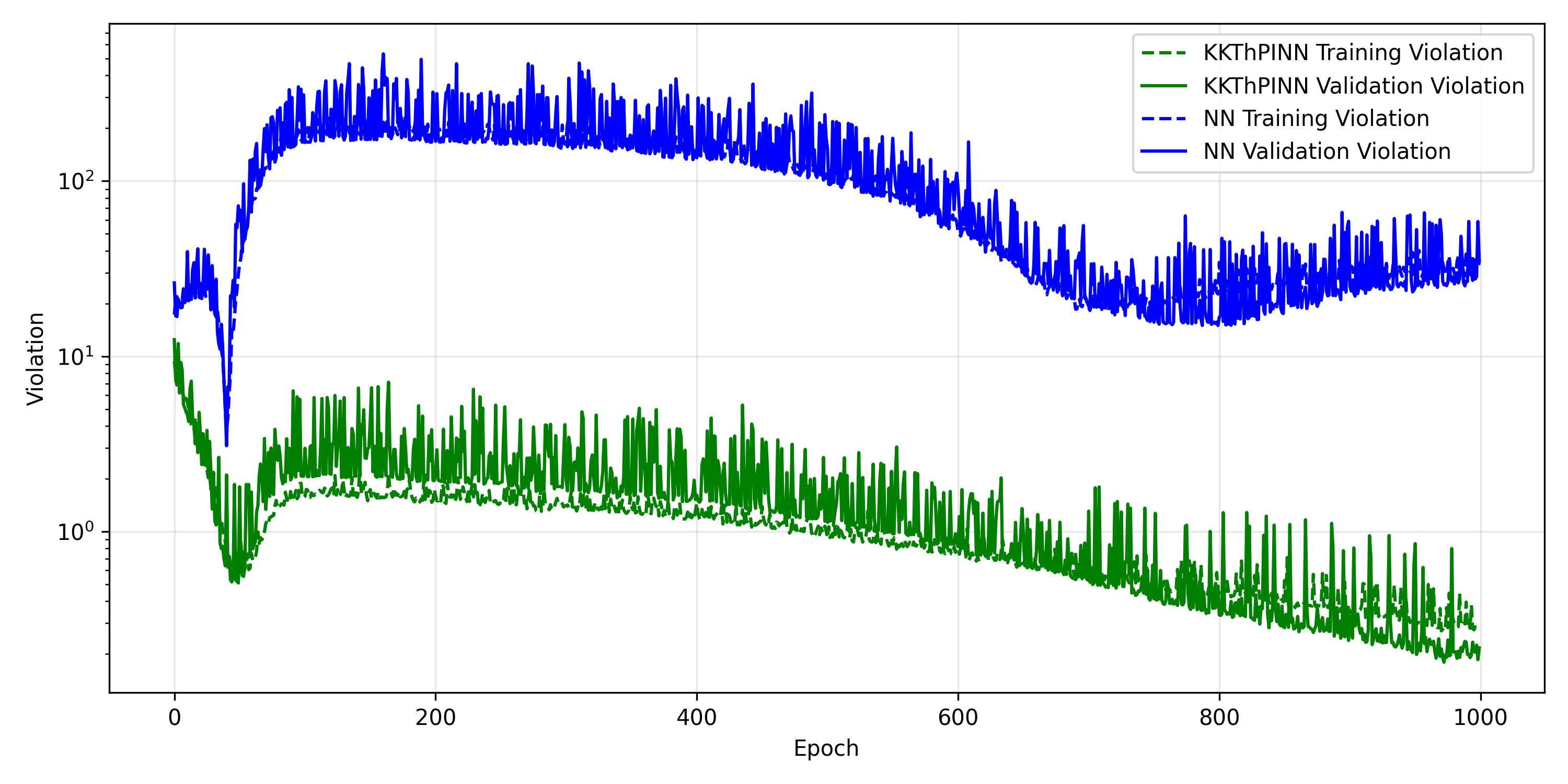}
        \caption{Nonlinear constraint violation}
        \label{fig:violation-2D}
    \end{subfigure}

    \caption{Training behavior of the standard neural network (NN) and the proposed PL-KKT-hPINN for the two-dimensional CSTR case, where \(T\) and \(C_{A0}\) are the inputs. The panels show: (a) training and validation RMSE over epochs, and (b) training and validation nonlinear constraint violation over epochs.}
    \label{fig:training_behavior_2D}
\end{figure}

Figures \ref{fig:training_behavior_1D} and \ref{fig:training_behavior_2D} show the loss function root-mean-square-error (RMSE) and mean constraint violation values during each epoch of training for the 1D and 2D cases, respectively. In both bases, the PL-KKT-hPINN converges faster than the standard NN with a nearly identical final RMSE. Moreover, the PL-KKT-hPINN decreases to by several orders-of-magnitude such that the mean constraint violation is the same magnitude of the mean piecewise-linear approximation errors. The approximation error can be further reduced by refining the input partition and increasing the number of linearization regions, as discussed in the Section~\ref{sec:segments}. Hence, these PL-KKT-hPINN is able to improve physical consistency without sacrificing predictive accuracy. 

\subsection{Effect of the Piecewise-Linear Approximation}
\label{sec:segments}

\begin{figure}[!htb]
    \centering

    \begin{subfigure}[t]{0.32\textwidth}
        \centering
        \includegraphics[width=\textwidth]{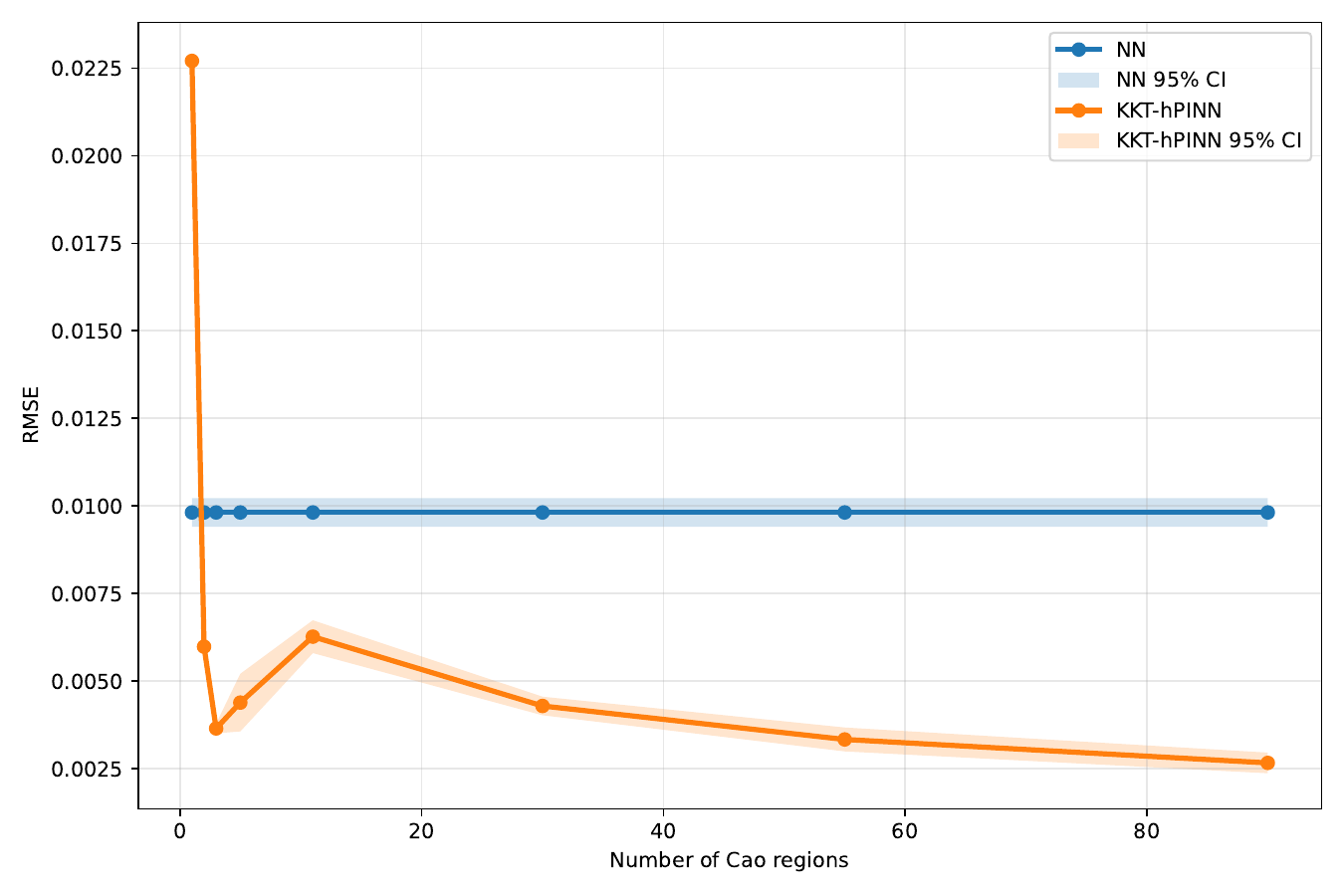}
        \caption{Test RMSE}
        \label{fig:rmse_segments1d}
    \end{subfigure}
    \hfill
    \begin{subfigure}[t]{0.32\textwidth}
        \centering
        \includegraphics[width=\textwidth]{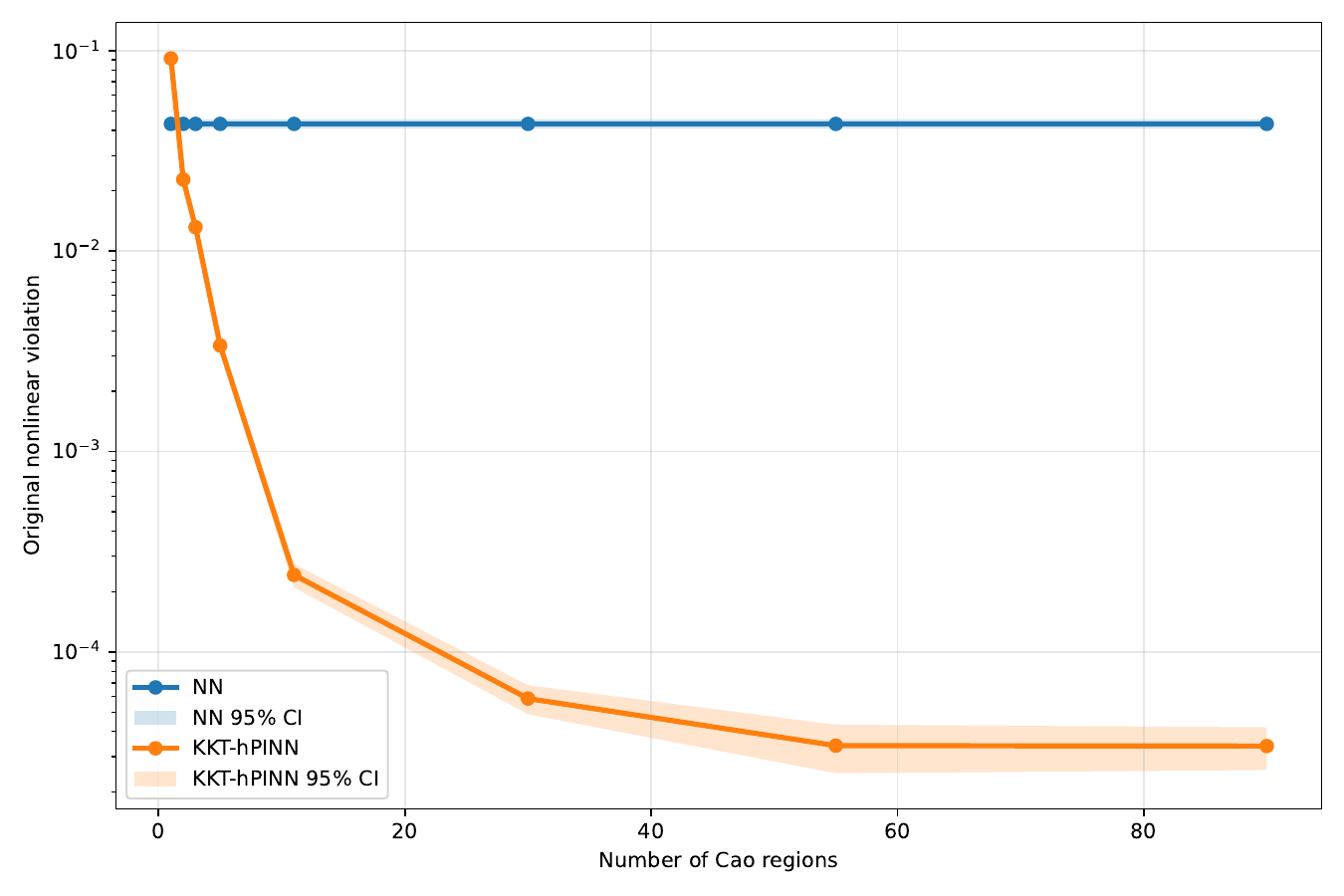}
        \caption{Nonlinear violation}
        \label{fig:viol_segments1d}
    \end{subfigure}
    \begin{subfigure}[t]{0.32\textwidth}
        \centering
        \includegraphics[width=\textwidth]{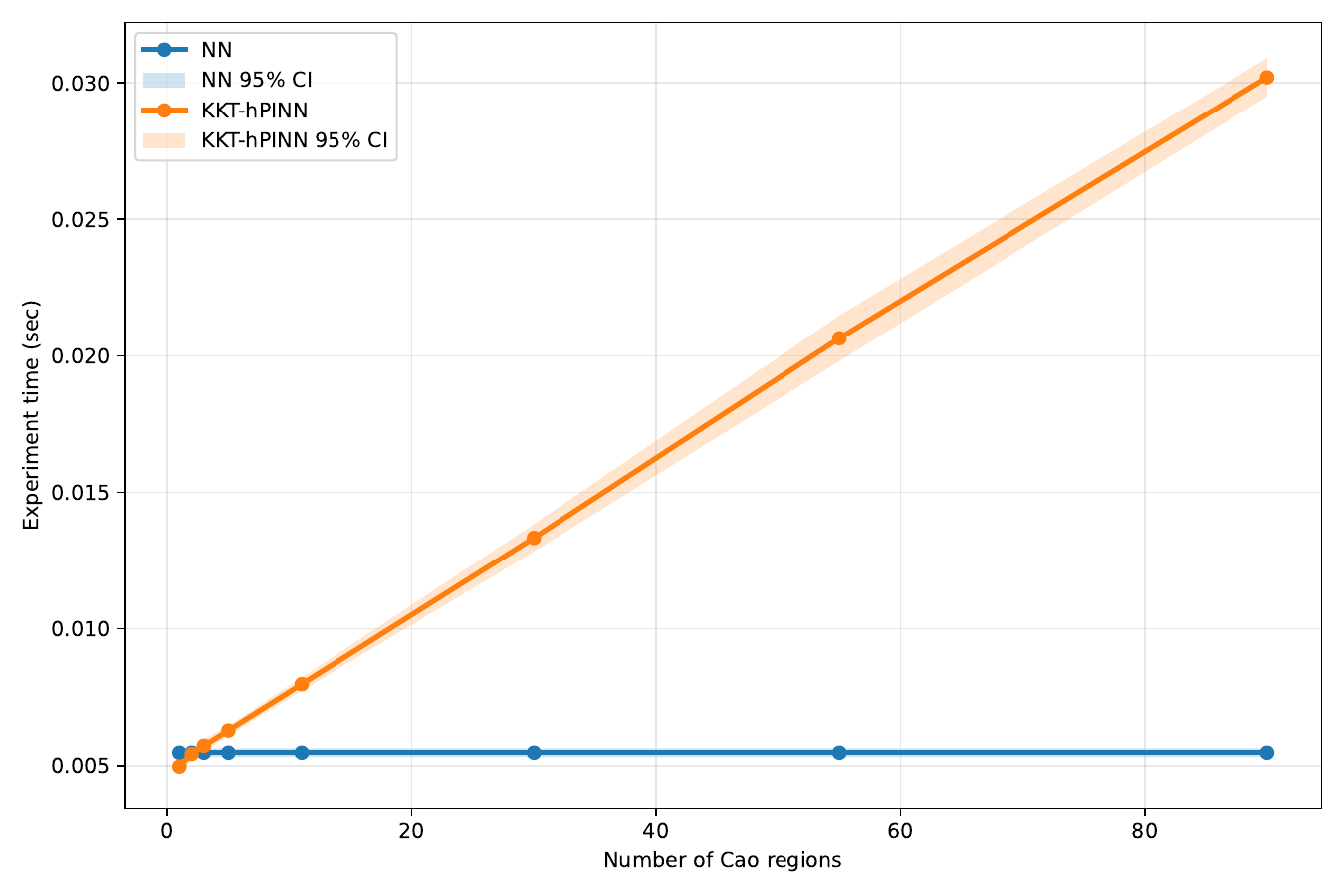}
        \caption{Experiment time}
        \label{fig:time_segments1d}
    \end{subfigure}

    \caption{Effect of the number of linearization regions on the performance of PL-KKT-hPINN for the one-dimensional CSTR case. The panels show: (a) test RMSE, (b) nonlinear constraint violation, and (c) experiment time. The standard NN is shown as a baseline because its architecture does not depend on the number of linearization regions, while PL-KKT-hPINN uses regional KKT projections based on the selected piecewise-linear partition. Markers represent the mean over 50 repeated runs, and shaded regions indicate 95\% confidence intervals.}
    \label{fig:segments_effect1d}
\end{figure}

\begin{figure}[!htb]
    \centering

    \begin{subfigure}[t]{0.32\textwidth}
        \centering
        \includegraphics[width=\textwidth]{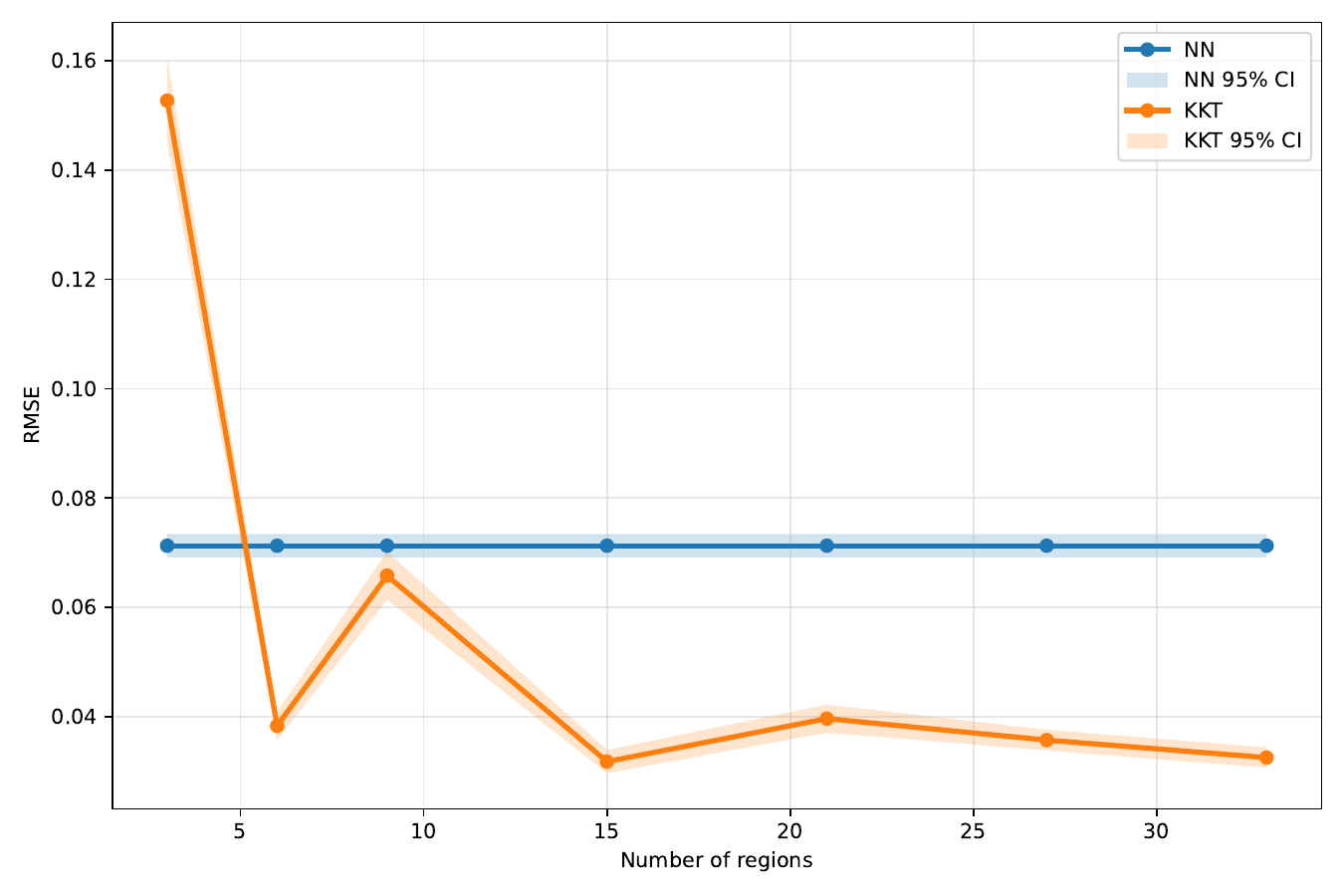}
        \caption{Test RMSE}
        \label{fig:rmse_segments}
    \end{subfigure}
    \hfill
    \begin{subfigure}[t]{0.32\textwidth}
        \centering
        \includegraphics[width=\textwidth]{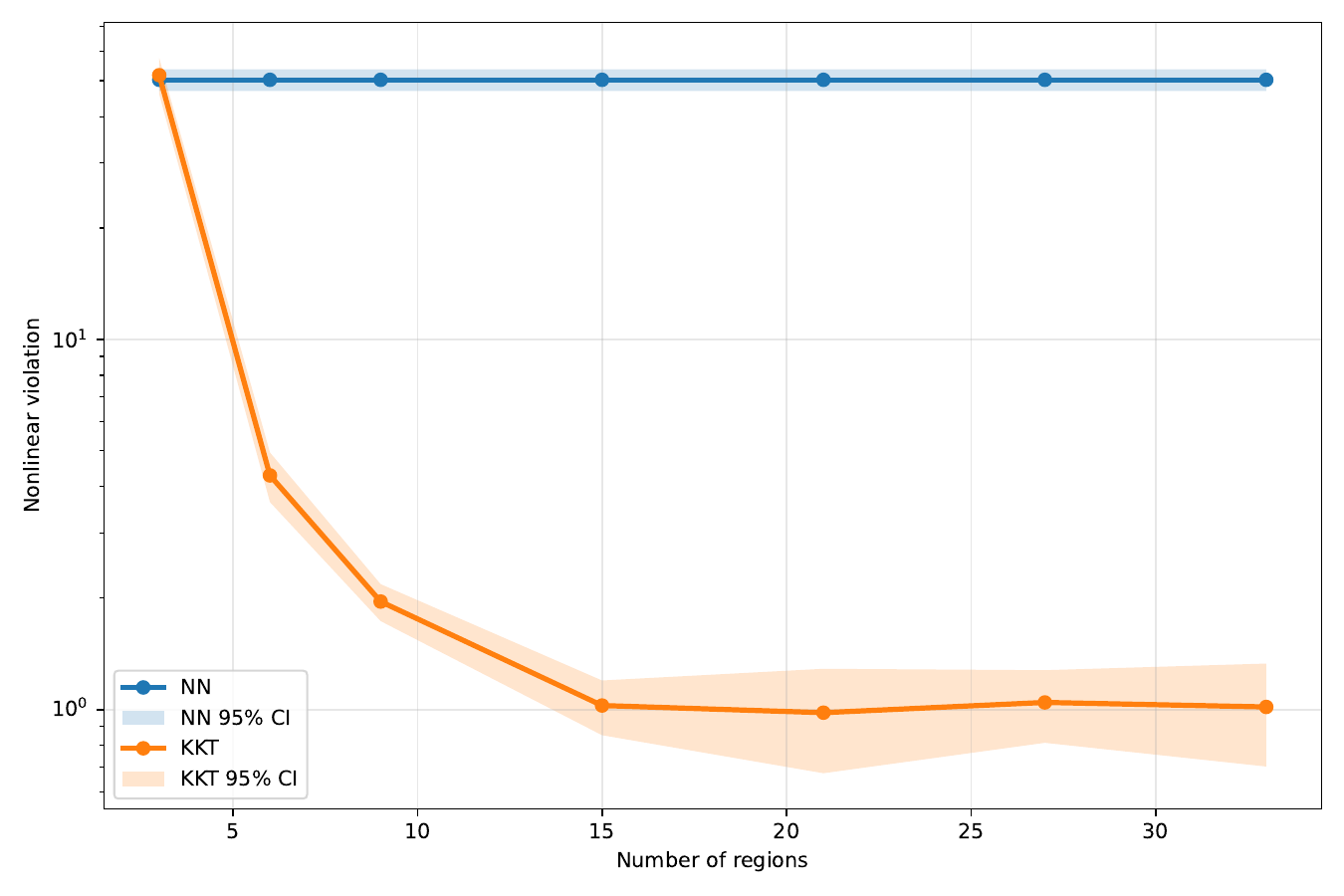}
        \caption{Nonlinear violation}
        \label{fig:viol_segments}
    \end{subfigure}
    \begin{subfigure}[t]{0.32\textwidth}
        \centering
        \includegraphics[width=\textwidth]{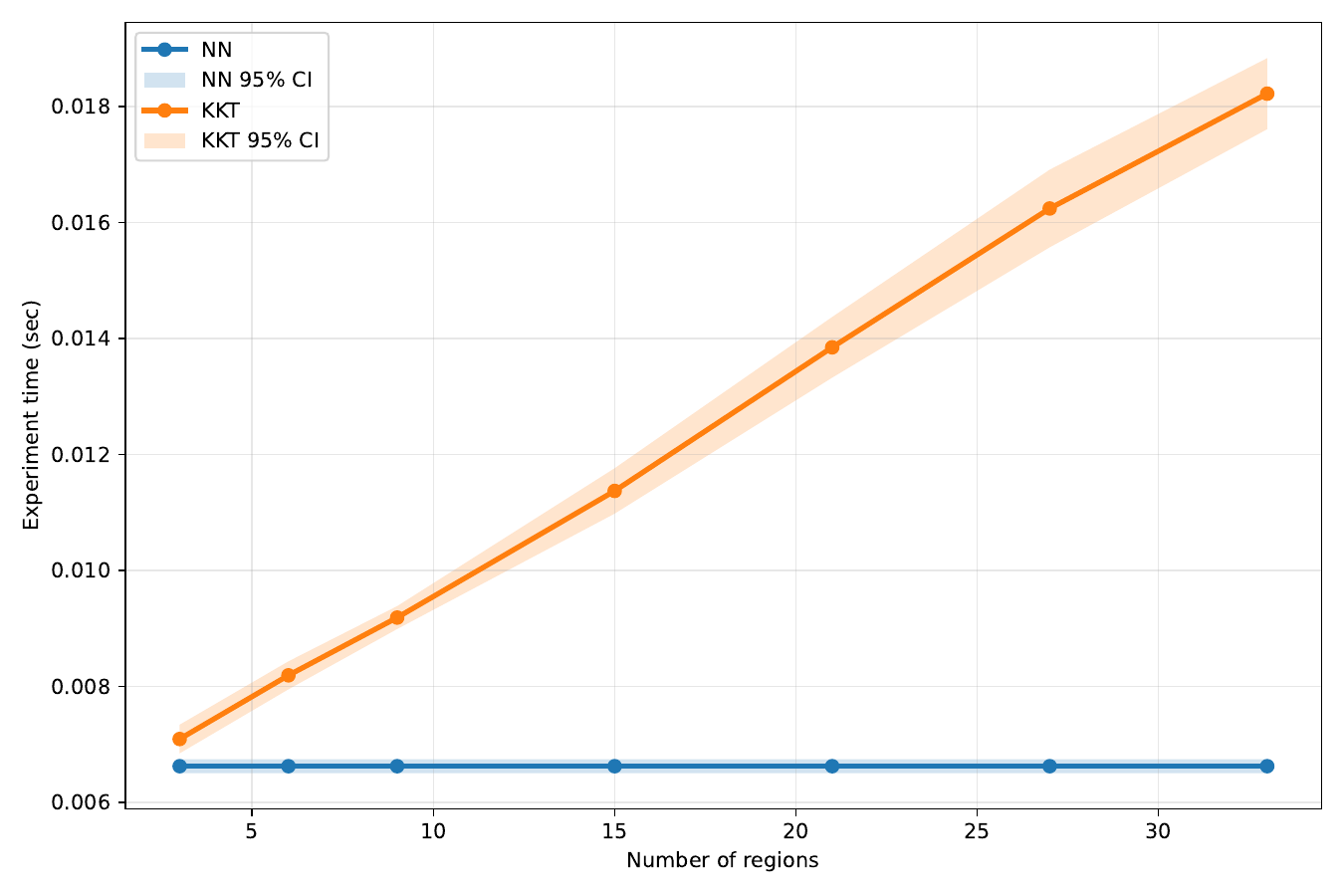}
        \caption{Experiment time}
        \label{fig:time_segments}
    \end{subfigure}

    \caption{Effect of the number of linearization regions on the performance of PL-KKT-hPINN for the two-dimensional CSTR case. The panels show: (a) test RMSE, (b) nonlinear constraint violation, and (c) experiment time. The standard NN is shown as a baseline because its architecture does not depend on the number of linearization regions, while PL-KKT-hPINN uses regional KKT projections based on the selected piecewise-linear partition. Markers represent the mean over 50 repeated runs, and shaded regions indicate 95\% confidence intervals.}
    \label{fig:segments_effect2d}
\end{figure}

To evaluate how the piecewise-linear approximation affects model performance, we vary the number of linearization regions with a fixed amount of training data. Figures \ref{fig:segments_effect1d} and \ref{fig:segments_effect2d} summarize the resulting loss function RMSEs, mean nonlinear constraint violations, and computational times at inference for both case studies. Moreover, the standard NN performance is provided as a baseline.

As expected, increasing the number of regions significantly decreases the mean constraint violation and even decreases the loss functions RMSE. Moreover, the mean constraint violations are of comparable magnitudes to the piecewise-linear approximation error over the training data which confirms that that the PL-KKT-hPINN does not introduce additional approximation error in enforcing constraints. This also means that the expected constraint satisfaction can be estimated based on the sampled approximation error of the piecewise-linear representation. We also observe that performance severely degrades when too few regions (e.g., two) are used, since the resulting piecewise-linear representation is too coarse to capture the behavior of the constraint.

Notably, the magnitude of constraint violation differs between the two cases with the 2D case exhibiting a larger violation. This behavior can be attributed to the Arrhenius terms inducing large derivatives with respect to temperature that make the local linearizations highly sensitive to small perturbations in $T$. This increased sensitivity limits the accuracy of the piecewise-linear representation which in turn leads to increases constraint violation as discussed in Section \ref{subsec:piecewise_linear_approximation}. In comparison, the 1D case study only depends on $C_{A0}$ as an input with fixed temperature which removes the influence of the Arrhenius terms on the gradients used in Equation \eqref{eq:taylor_scalar_constraint}. 

Finally, we observe that increasing the number of regions moderately increases the computational cost at inference. The linear scaling is expected as discussed in Section \ref{subsec:piecewise_linear_approximation}; moreover, the implementation used does not exploit the parallel structure of the regional projections which means the computational cost can be decreased further. Overall, these results show that increasing the number of linearization regions provides a systematic way to improve nonlinear constraint enforcement, while preserving predictive accuracy and maintaining a static, non-iterative architecture during training and inference.

\subsection{Data Efficiency}

\begin{figure}[!htb]
    \centering

    \begin{subfigure}[t]{0.48\textwidth}
        \centering
        \includegraphics[width=\linewidth,height=5.2cm]{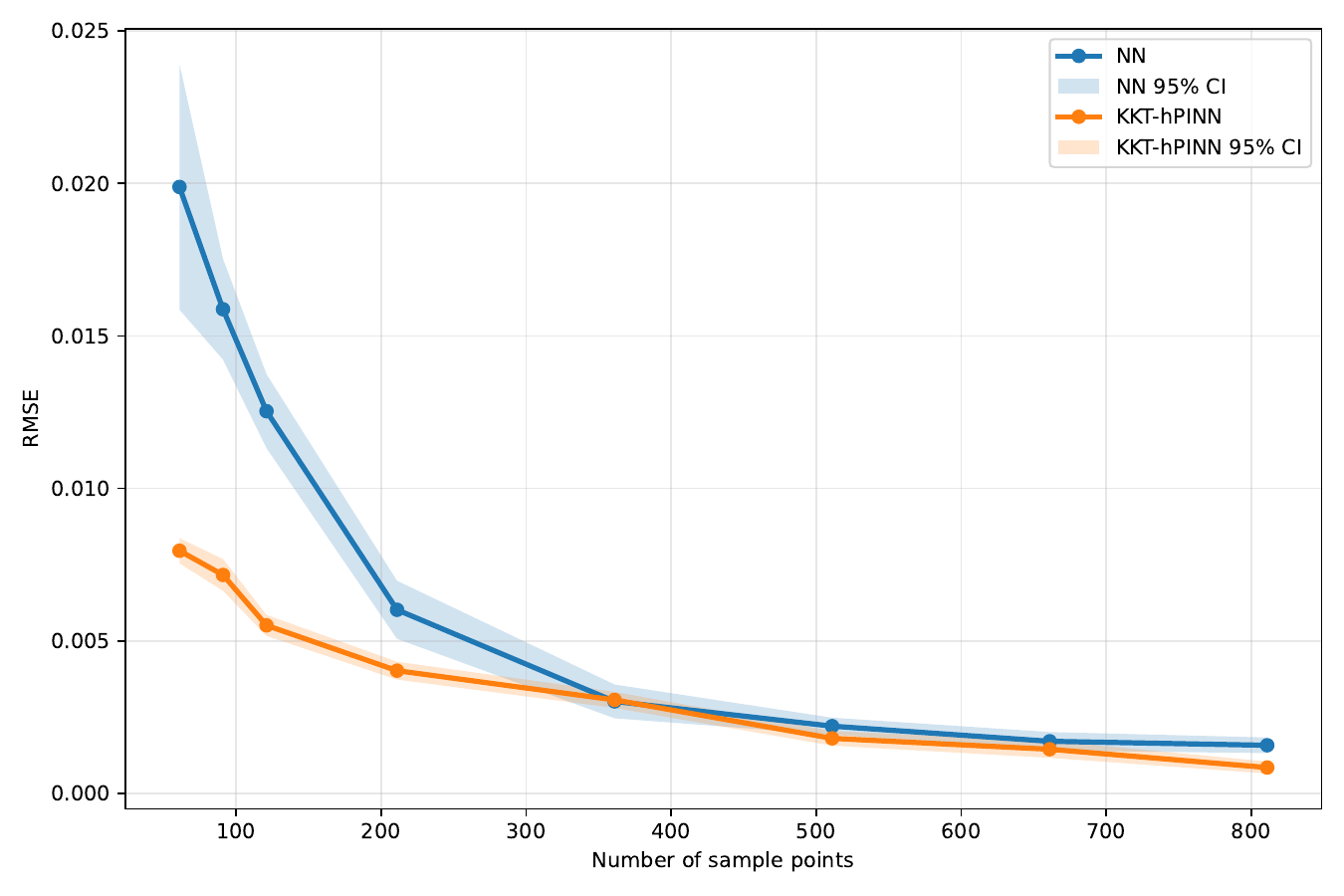}
        \caption{One-dimensional input case}
        \label{fig:RMSE1D}
    \end{subfigure}
    \hfill
    \begin{subfigure}[t]{0.48\textwidth}
        \centering
        \includegraphics[width=\linewidth,height=5.2cm]{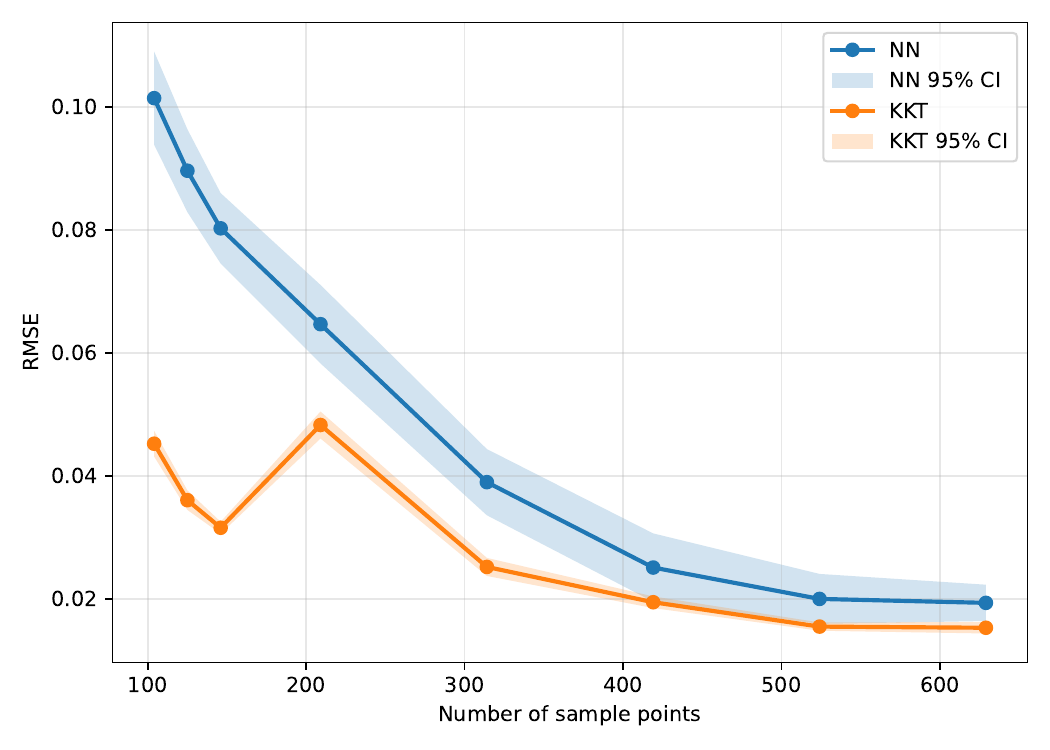}
        \caption{Two-dimensional input case}
        \label{fig:RMSE2D}
    \end{subfigure}

    \caption{Effect of training sample size on test RMSE for the CSTR case study. Panel (a) corresponds to the one-dimensional input case, where inlet concentration $C_{A0}$ is the input. Panel (b) corresponds to the two-dimensional input case, where temperature $T$ and feed concentration $C_{A0}$ are used as inputs. In both cases, the standard NN and PL-KKT-hPINN are trained using different numbers of sample points to evaluate data efficiency. Markers represent the mean over 50 repeated runs, and shaded regions indicate 95\% confidence intervals.}
    \label{fig:RMSE_data_efficiency}
\end{figure}

We assess how robust PL-KKT-hPINN is to overfitting with limited training data relative to a standard NN. Figure \ref{fig:RMSE_data_efficiency} summarizes the loss function RMSE of these models after they are training with datasets of decreasing size. In both both cases, the RMSE increases with decreased dataset cardinality. However, the advantage of PL-KKT-hPINN is more evident in the low-data regime where its improved robustness can likely be attributed to the embedded physical constraint that restricts the admissible prediction space, reducing overfitting when training data is limited. This highlights how the constraint projection acts to regularize the NN making it more robust to overfitting \cite{zavala2025}. 

\subsection{Comparison to PINNs}

\begin{figure}[!htp]
    \centering
    \includegraphics[width=0.5\textwidth]{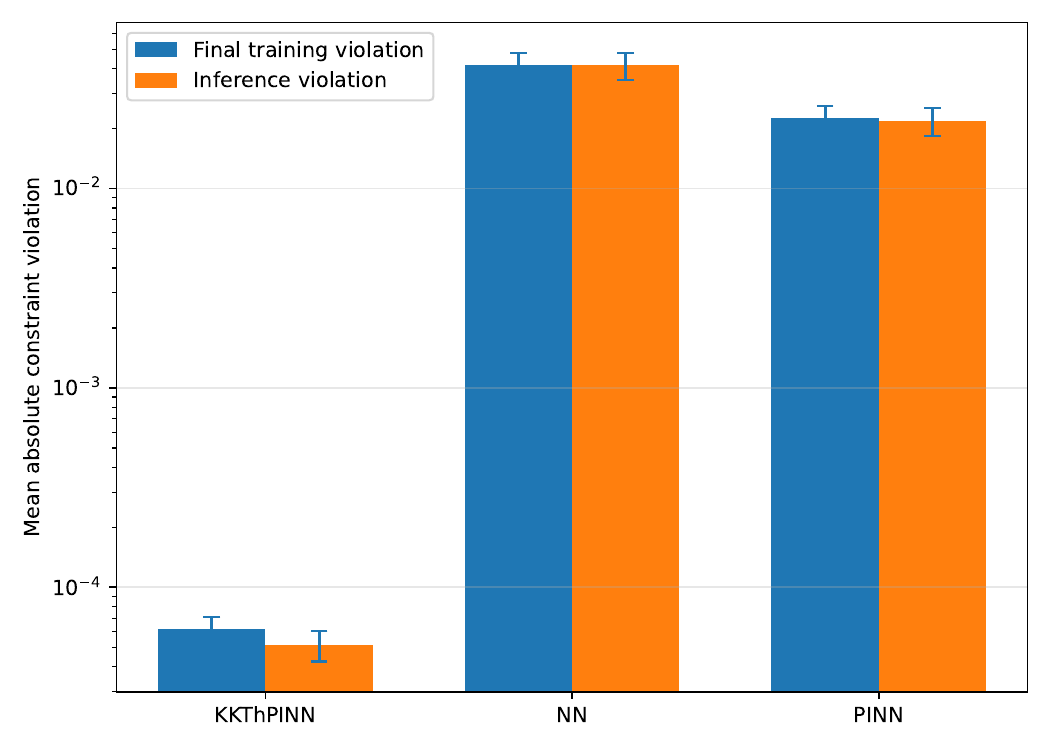}
    \caption{Comparison of the final mean absolute nonlinear constraint violation during training and inference for the standard NN, PINN, and PL-KKT-hPINN. The PINN results correspond to the selected penalty weight \(\mu=[0.01 \; 0.05]^T\). Error bars indicate variability across repeated runs. PL-KKT-hPINN achieves the lowest violation in both training and inference, while PINN improves over the standard NN but remains less accurate in constraint satisfaction than PL-KKT-hPINN.}
    \label{fig:pinn_train_infer_violation}
\end{figure}

To further benchmark the proposed method, we juxtapose the performance of the PL-KKT-hPINN with a soft-constrained PINN. During training, Equation \eqref{eq:f_nonlin} added to loss function as a weighted penalty term:
\[
\mathcal{L}_{\mathrm{PINN}}(\mathbf{x}, \mathbf{y}; \Theta) = \mathcal{L}_{\mathrm{NN}}(\mathbf{x}, \mathbf{y}; \Theta) + \mu_1 \cdot g_1(\mathbf{x}, \mathbf{y}) + \mu_2 \cdot g_2(\mathbf{x}, \mathbf{y}),
\]
$\mu_i\in [0, \infty)$ are weighting parameters that balance prediction error and constraint violation. Using a Pareto frontier constructed of 30 candidate $\mu$ values, we select $\mu = [0.01 \; 0.05]^T$ as the point that minimizes constraint violations. We compare the PINN in terms of constraint violation against a PL-KKT-hPINN with 30 regions and a standard NN; Figure \ref{fig:pinn_train_infer_violation} summarizes the results. The PINN modestly reduces the nonlinear violation relative to the NN, showing that the added physics penalty improves physical consistency; however, its violation remains orders-of-magnitude larger than that of PL-KKT-hPINN. Moreover, all models exhibit nearly identical prediction errors. This comparison highlights the practical advantages of PL-KKT-hPINN over soft-constrained PINNs in that the former guarantee constraint enforcement (during training and inference) and avoid the need for parameter tuning (eliminating the need to repeatedly retrain).

\section{Conclusions and Future Work}
\label{sec:conclusions}

In this study, we propose PL-KKT-hPINN as an extension to KKT-hPINN to enforce nonlinear equality constraints via projection on piecewise-linear approximations. Unlike other approaches in the literature, this provides a static NN architecture that is noniterative and readily amendable to parallelization. The approximation error induced by this method is solely introduced by the piecewise-linear representation which can be readily quantified apriori using classical techniques. Case studies demonstrate that PL-KKT-hPINNs can outperform standard NNs and PINNs in terms of constraint satisfaction (with guarantees), prediction accuracy, and data efficiency. Moreover, PL-KKT-hPINNs avoid the challenge of tuning weighting parameters in PINNs. All this make PL-KKT-hPINN a promising approach for surrogate modeling of nonlinear chemical engineering systems.

Worthwhile directions of future work include investigating the use of alternative piecewise-linear representations within the framework such as those provided by linear decision trees. Such an extension would also motivate the investigation of efficient ways to formulate the activation functions for regions defined by hierarchical rules. Finally, the utility of parallelization should be further explored how this can further reduce the computation cost imposed by increasing the number of linearization regions.

\section*{Acknowledgments}
This work was supported by Natural Sciences and Engineering Research Council of Canada under grant  50503-10882. 

\bibliography{refs}

@article{mcbride2019overview,
  title={Overview of surrogate modeling in chemical process engineering},
  author={McBride, Kevin and Sundmacher, Kai},
  journal={Chemie Ingenieur Technik},
  volume={91},
  number={3},
  pages={228--239},
  year={2019},
  publisher={Wiley Online Library}
}

@article{misener2023formulating,
  title={Formulating data-driven surrogate models for process optimization},
  author={Misener, Ruth and Biegler, Lorenz},
  journal={Computers \& Chemical Engineering},
  volume={179},
  pages={108411},
  year={2023},
  publisher={Elsevier}
}

@article{bhosekar2018advances,
  title={Advances in surrogate based modeling, feasibility analysis, and optimization: A review},
  author={Bhosekar, Atharv and Ierapetritou, Marianthi},
  journal={Computers \& Chemical Engineering},
  volume={108},
  pages={250--267},
  year={2018},
  publisher={Elsevier}
}

@article{bradley2022perspectives,
  title={Perspectives on the integration between first-principles and data-driven modeling},
  author={Bradley, William and Kim, Jinhyeun and Kilwein, Zachary and Blakely, Logan and Eydenberg, Michael and Jalvin, Jordan and Laird, Carl and Boukouvala, Fani},
  journal={Computers \& Chemical Engineering},
  volume={166},
  pages={107898},
  year={2022},
  publisher={Elsevier}
}

@article{williams2021selection,
  title={Selection of surrogate modeling techniques for surface approximation and surrogate-based optimization},
  author={Williams, Bianca and Cremaschi, Selen},
  journal={Chemical Engineering Research and Design},
  volume={170},
  pages={76--89},
  year={2021},
  publisher={Elsevier}
}

@article{chen2024physics,
  title={Physics-informed neural networks with hard linear equality constraints},
  author={Chen, Hao and Flores, Gonzalo E Constante and Li, Can},
  journal={Computers \& Chemical Engineering},
  volume={189},
  pages={108764},
  year={2024},
  publisher={Elsevier}
}

@article{cybenko1989approximation,
  title={Approximation by superpositions of a sigmoidal function},
  author={Cybenko, George},
  journal={Mathematics of control, signals and systems},
  volume={2},
  number={4},
  pages={303--314},
  year={1989},
  publisher={Springer}
}

@article{hornik1989multilayer,
  title={Multilayer feedforward networks are universal approximators},
  author={Hornik, Kurt and Stinchcombe, Maxwell and White, Halbert},
  journal={Neural networks},
  volume={2},
  number={5},
  pages={359--366},
  year={1989},
  publisher={Elsevier}
}

@article{kim2020surrogate,
  title={Surrogate-based optimization for mixed-integer nonlinear problems},
  author={Kim, Sun Hye and Boukouvala, Fani},
  journal={Computers \& Chemical Engineering},
  volume={140},
  pages={106847},
  year={2020},
  publisher={Elsevier}
}

@article{dias2020integration,
  title={Integration of planning, scheduling and control problems using data-driven feasibility analysis and surrogate models},
  author={Dias, Lisia S and Ierapetritou, Marianthi G},
  journal={Computers \& Chemical Engineering},
  volume={134},
  pages={106714},
  year={2020},
  publisher={Elsevier}
}

@incollection{mohammadi2022surrogate,
  title={Surrogate modeling and surrogate-based optimization with stochastic simulations},
  author={Mohammadi, Samira and Williams, Bianca and Cremaschi, Selen},
  booktitle={Computer Aided Chemical Engineering},
  volume={49},
  pages={31--40},
  year={2022},
  publisher={Elsevier}
}

@article{raissi2019physics,
  title={Physics-informed neural networks: A deep learning framework for solving forward and inverse problems involving nonlinear partial differential equations},
  author={Raissi, Maziar and Perdikaris, Paris and Karniadakis, George E},
  journal={Journal of Computational physics},
  volume={378},
  pages={686--707},
  year={2019},
  publisher={Elsevier}
}

@article{ma2022data,
  title={Data-driven strategies for optimization of integrated chemical plants},
  author={Ma, Kaiwen and Sahinidis, Nikolaos V and Amaran, Satyajith and Bindlish, Rahul and Bury, Scott J and Griffith, Devin and Rajagopalan, Sreekanth},
  journal={Computers \& Chemical Engineering},
  volume={166},
  pages={107961},
  year={2022},
  publisher={Elsevier}
}

@article{beucler2021enforcing,
  title={Enforcing analytic constraints in neural networks emulating physical systems},
  author={Beucler, Tom and Pritchard, Michael and Rasp, Stephan and Ott, Jordan and Baldi, Pierre and Gentine, Pierre},
  journal={Physical review letters},
  volume={126},
  number={9},
  pages={098302},
  year={2021},
  publisher={APS}
}

@inproceedings{amos2017optnet,
  title={Optnet: Differentiable optimization as a layer in neural networks},
  author={Amos, Brandon and Kolter, J Zico},
  booktitle={International conference on machine learning},
  pages={136--145},
  year={2017},
  organization={PMLR}
}

@article{agrawal2019differentiable,
  title={Differentiable convex optimization layers},
  author={Agrawal, Akshay and Amos, Brandon and Barratt, Shane and Boyd, Stephen and Diamond, Steven and Kolter, J Zico},
  journal={Advances in neural information processing systems},
  volume={32},
  year={2019}
}

@article{min2024hard,
  title={Hard-constrained neural networks with universal approximation guarantees},
  author={Min, Youngjae and Azizan, Navid},
  journal={arXiv preprint arXiv:2410.10807},
  year={2024}
}

@article{mukherjee2024development,
  title={Development of steady-state and dynamic mass and energy constrained neural networks for distributed chemical systems using noisy transient data},
  author={Mukherjee, Angan and Bhattacharyya, Debangsu},
  journal={Industrial \& Engineering Chemistry Research},
  volume={63},
  number={32},
  pages={14211--14239},
  year={2024},
  publisher={ACS Publications}
}

@article{lastrucci2025picard,
  title={Picard-KKT-hPINN: Enforcing Nonlinear Enthalpy Balances for Physically Consistent Neural Networks},
  author={Lastrucci, Giacomo and Karia, Tanuj and Gromotka, Zo{\"e} and Schweidtmann, Artur M},
  journal={arXiv preprint arXiv:2501.17782},
  year={2025}
}

@article{lastrucci2025enforce,
  title={ENFORCE: Nonlinear Constrained Learning with Adaptive-depth Neural Projection},
  author={Lastrucci, Giacomo and Schweidtmann, Artur M},
  journal={arXiv preprint arXiv:2502.06774},
  year={2025}
}

@article{iftakher2025physics,
  title={Physics-informed neural networks with hard nonlinear equality and inequality constraints},
  author={Iftakher, Ashfaq and Golder, Rahul and Roy, Bimol Nath and Hasan, MM Faruque},
  journal={Computers \& Chemical Engineering},
  pages={109418},
  year={2025},
  publisher={Elsevier}
}

@article{misener2010piecewise,
  title={Piecewise-linear approximations of multidimensional functions},
  author={Misener, R and Floudas, CA2602908},
  journal={Journal of optimization theory and applications},
  volume={145},
  number={1},
  pages={120--147},
  year={2010},
  publisher={Springer}
}

@article{ammari2023linear,
  title={Linear model decision trees as surrogates in optimization of engineering applications},
  author={Ammari, Bashar L and Johnson, Emma S and Stinchfield, Georgia and Kim, Taehun and Bynum, Michael and Hart, William E and Pulsipher, Joshua and Laird, Carl D},
  journal={Computers \& Chemical Engineering},
  volume={178},
  pages={108347},
  year={2023},
  publisher={Elsevier}
}

@article{mukherjee2026physics,
  title={Physics-constrained machine learning for chemical engineering},
  author={Mukherjee, Angan and Zavala, Victor M},
  journal={Current Opinion in Chemical Engineering},
  volume={51},
  pages={101228},
  year={2026},
  publisher={Elsevier}
}

@article{ralph1997sensitivity,
  title={Sensitivity analysis of composite piecewise smooth equations},
  author={Ralph, Daniel and Scholtes, Stefan},
  journal={Mathematical Programming},
  volume={76},
  number={3},
  pages={593--612},
  year={1997},
  publisher={Springer}
}

@Book{zavala2025,
  title     = {Statistics for Chemical Engineers},
  author    = {Victor M. Zavala},
  publisher = {Cambridge University Press},
  year      = {2025}
}
\end{document}